%% file: iclr2027_conference.tex
\documentclass{article} 
\usepackage{iclr2027_conference,times}
\usepackage{graphicx}
\input{math_commands.tex}

\usepackage{amsmath}
\usepackage{hyperref}
\usepackage{url}
\usepackage{caption}

\newcommand{\threlu}{\operatorname{ReLU}}
\newcommand{\throt}{\operatorname{Rot}}
\newcommand{\thlip}{\operatorname{Lip}}
\newcommand{\thind}{\mathbf{1}}
\newtheorem{theorytheorem}{Theorem}[section]
\newtheorem{theorylemma}[theorytheorem]{Lemma}
\newtheorem{theorycorollary}[theorytheorem]{Corollary}

\newenvironment{theoryproof}[1][Proof]{\par\noindent\textit{#1.}\ }%
  {\unskip\nobreak\hfill\rule{0.6em}{0.6em}\par\medskip}

\title{Capability and Interpretability of Transformers on Multi-task Recurrence}

\author{
  \makebox[\dimexpr\textwidth-2\tabcolsep\relax][c]{%
\begin{tabular}{@{}c@{}}
 \\
      Xiamin Chen$^{1}$, Linyu Liu$^{2}$,
      Jiachen Hu$^{1}$, Pinyan Lu$^{1,2}$\\[0.4em]
      \normalfont $^{1}$Shanghai University of Finance and Economics\\[0.1em]
      \normalfont $^{2}$Taylor Lab, Huawei Technologies Ltd.
    \end{tabular}%
  }
}

\usepackage{xcolor}

\iclrfinalcopy 
\begin{document}

\maketitle
\vspace{-1em}
\fancyhead[L]{Preprint}
\begin{abstract}
Transformers are widely used in various language tasks, with limited understanding of their internal mechanisms. Motivated by the recurrent structure of many linguistic rules, we study the capability and interpretability of Transformers through a controlled family of linear recurrence tasks over finite cyclic groups. These tasks range from executing a single fixed rule to identifying a latent rule from context and switching between explicitly specified rules, with and without input masking. Through extensive experiments across different configurations, we find that switching specified rules is simpler than identifying latent rules for Transformers in general. We also identify how several key factors affect the generalization capability of Transformers, including the depth, width, and the algebraic structure of the linear recurrence. Curriculum training is extremely useful when the number of rules is large.
With standard tools in mechanistic interpretability like linear probing and input patching, our study reveals how Transformers represent, identify, and execute multiple rules. 

\end{abstract}

\input{intro}

\input{problem}

\input{capability}

\input{mechanism}

\input{conclusion}
\input{AI_use_statement}

\bibliography{iclr2027_conference}
\bibliographystyle{iclr2027_conference}

\appendix
\newpage
\input{appendix}

\input{theory}

\end{document}

%% file: math_commands.tex
\usepackage{amsmath,amsfonts,bm}

\def\1{\bm{1}}

\DeclareMathAlphabet{\mathsfit}{\encodingdefault}{\sfdefault}{m}{sl}
\SetMathAlphabet{\mathsfit}{bold}{\encodingdefault}{\sfdefault}{bx}{n}

\DeclareMathOperator*{\argmax}{arg\,max}

%% file: intro.tex
\section{Introduction}
Natural language often exhibits strong local statistical properties. Syntactic dependencies tend to be short across
languages \citep{futrell2015dependency}, and classical $n$-gram models exploit
restricted histories for prediction \citep{brown1992class}. Nevertheless, applying a
basic simple rule can require selecting the appropriate context. For example, in sentence ``The keys
to the cabinet are on the table,'' the verb agrees with the subject head
\emph{keys}, rather than the nearest noun \emph{cabinet}
\citep{linzen2016syntax}. More generally, natural language is equipped with
systematic reuse of constituents and combination of basic sequence rules
\citep{hupkes2020compositionality}. These observations motivate controlled
investigations of how language encodes multiple basic rules.

Transformers have achieved strong performance in machine translation
\citep{vaswani2017attention}, language modeling \citep{brown2020fewshot},
visual recognition \citep{dosovitskiy2021image}, and sequential
decision-making \citep{chen2021decision}. Although these tasks consist of the compositions of many basic rules, it remains unclear how Transformers learn and generalize multiple basic rules. 
We therefore ask:
\begin{center}
\emph{How and when do
Transformers learn multiple basic rules? What internal representations
and mechanisms support their predictions?}    
\end{center}

We investigate these questions through sequences generated by $q_t=f_{a_t}(q_{t-1},\ldots,q_{t-k})$, where $a_t$ is a basic rule. Specifically, we focus on linear recurrences
\[
q_t=\sum_{i=1}^{k}(a_t)_i q_{t-i}\pmod p,
\qquad q_t\in\mathbb F_p,\quad
 a_t\in\mathcal A\subseteq\mathbb F_p^k,
\]
where $k$ is the order of the recurrence and $\mathcal A$ is a pool
of $N$ rules. This family provides a suitable interface to answer the questions above: the generating rules are explicit, task difficulty can be varied systematically, and candidate prediction strategies can be tested conveniently.

We study two possible combinations of multiple rules: the \emph{inter-trajectory} setting and the \emph{intra-trajectory} setting.
In the inter-trajectory setting, each sequence is determined by a single unobserved rule, requiring the model to infer predictive information about that rule from context. In the intra-trajectory setting, the rule can change at every step and is specified by an input tag, requiring conditional execution of successive rules. These two settings are mainly motivated from the fact that natural language prediction requires both adaptation to persistent contextual regularities and flexible application of grammatical operations. Readers must adjust their syntactic expectations to the distributions encountered in a linguistic environment \citep{fine2013adaptation}, or select different realizations of the grammar within the same sentence \citep{linzen2016syntax}.

Motivated by language processing with partially unavailable context
\citep{futrell2020lossy}, we also study the effect of masking, which may remove operands of the immediate
update and require prediction from other observed values. We vary rule count, recurrence order, and maximum
mask length across model depths and widths, and combine training
interventions, rule probes, and input perturbations to investigate underlying mechanism. Our main findings and contributions can be summarized as follows:

\begin{itemize}
\item \emph{Intra-trajectory setting is easier than inter-trajectory setting}: With multiple rules, Transformers generalize more easily when the active rule is explicitly provided than when it must be inferred from context. As the number of rules increases, we also find that curriculum learning helps to learn better than joint training.

\item \emph{Depth is crucial for handling multi-rule masked observations}: Greater depth improves prediction as mask length grows in order to infer the masked contents. Interestingly, we also find that masking turns a single-rule problem into a multi-rule one by forcing the model to predict with different mask patterns. 

\item \emph{Algebraic structure matters in high order tasks}:
Some higher-order rules admit equivalent recurrences using fewer operands as a \emph{shortcut solution}
at longer lags.Adjusting the weights of early positions helps
control the learning of these shortcuts.

\item \emph{Revealing underlying mechanism of Transformers}: We reveal how Transformers represent and execute multiple rules with attention analysis, linear probing and input patching. Moreover, we also find how data ratio across tasks affect the performance of the networks.

\end{itemize}

The remainder of this paper is organized as follows. Section~\ref{sec:setup} introduces the recurrence tasks and experimental setup. Section~\ref{sec:model_training} studies the generalization capability across different tasks and model sizes. Section~\ref{sec:curriculum} examines how curriculum learning can enhance this capability. Section~\ref{sec:rule_mechanisms} investigates how Transformers learn, represent, and execute multiple rules. Section~\ref{sec:conclusion} concludes the paper. Due to space limits, we defer related work to Appendix~\ref{sec:related}.

%% file: problem.tex
\section{recurrence Sequence: Problem and Experimental Setup}
\label{sec:setup}

\subsection{Problem Setup}

We study recurrence sequences in which each new element is determined by previous elements and a fixed rule. We first consider a length-$T$ sequence determined by two initial values $(q_1,q_2)$ and a rule $a=(a_1,a_2)$:
\begin{equation}
q_t=f_a(q_{t-2},q_{t-1})=
a_1q_{t-1}+a_2q_{t-2}
\pmod p,
\qquad t=3,\ldots,T,
\label{equ:basic_rule}
\end{equation}
where $q_t\in \mathbb{F}_p,a\in \mathbb{F}_p^2\setminus{(0,0)}$, and $\mathbb{F}_p=\mathbb{Z}/p\mathbb{Z}$. Given $(q_1,q_2)$ and $a$, all later elements are uniquely determined by Eq.~\eqref{equ:basic_rule}.

This recurrence family includes the modular addition task commonly used to study \textit{grokking} \citep{power2022grokking,nanda2023progress} as a special case: setting $a=(1,1)$ and $T=3$ gives $q_3=(q_1+q_2)\bmod p$. Our setup extends this case from a single-step prediction to sequence prediction, which allows us to study how Transformers \citep{vaswani2017attention} learn and use rules over sequences. We further extend it by varying the recurrence order and how multiple rules are organized:

\begin{itemize}
\item \textbf{Order-$k$ single-rule setting.}
We first extend the order-$2$ rule in Eq.~\eqref{equ:basic_rule} to order $k$. A fixed rule $a\in \mathbb{F}_p^k$ determines each new element from the previous $k$ elements:
$
q_t
=
f_a(q_{t-k},\ldots,q_{t-1})
=
\sum_{i=1}^{k}a_iq_{t-i}
\pmod p.
$
This setting tests whether the model can learn a rule that depends on a longer history.

\item \textbf{Inter-trajectory multi-rule setting.}
We next consider a fixed pool of $N$ distinct recurrence rules,
$\mathcal{A} = \{a^{(1)}, \ldots, a^{(N)}\}.
$
For each rule $a^{(j)}$, we generate a dataset
$\mathcal{D}_j$ consisting of sequences satisfying
$
q_t = f_{a^{(j)}}(q_{t-2},q_{t-1}),
\  t=3,\ldots,T.
$
The training data are constructed separately for each
rule and then combined and shuffled:
$
\mathcal{D}_{\mathrm{train}}
= \operatorname{Shuffle}\left(
\biguplus_{j=1}^{N} \mathcal{D}_j^{\mathrm{train}}
\right).
$
Thus, each sequence follows a single rule, while
the model is jointly trained on sequences from all
$N$ rules. Since the rule identity is not provided,
the model must infer the underlying rule from
the observed sequence to predict the next element.

\item \textbf{Intra-trajectory multi-rule setting.}
We also allow the rule to change within a sequence. The input takes the form
$
[q_1,q_2,c_3,q_3,c_4,q_4,c_5,q_5,\ldots],
$
where each rule tag $c_t$ is sampled from a fixed rule pool $\mathcal{A}$ and tells the model which rule to use at step $t$. Let $\hat{c}_t \in \mathcal{A}$ denote the rule represented by $c_t$. Then
$
q_t=f_{\hat{c}_t}(q_{t-2},q_{t-1}).
$
Unlike the inter-trajectory setting, the model does not need to infer which rule is active from earlier values alone, since $c_t$ gives this information explicitly. Instead, it must read the current rule tag and apply the corresponding rule to the previous elements. This setting provides a simple model of sequences whose local pattern can change over time.

\end{itemize}

We further increase task difficulty through \textit{masking},
which introduces missing observations into the sequence.
Let $m_t \in \{0,1\}$ indicate whether the value at
position $t$ is masked (we do not mask rule tags). The observed sequence
$\tilde{q}=(\tilde{q}_1,\ldots,\tilde{q}_T)$ is defined by
\[
\tilde{q}_t =
\begin{cases}
q_t, & m_t=0,\\
\bot, & m_t=1,
\end{cases}
\]
where $\bot$ is a special mask token and the original
positions are preserved.

For example, under the rule
$q_t=q_{t-1}+q_{t-2}\pmod{127}$,
a sequence may be observed as
\[
\tilde{q} =
[1,2,3,\bot,\bot,13,21,34,\ldots].
\]
Without masking, an order-$2$ recurrence requires only
the two most recent values. When these values are
masked, however, the model may need to retrieve
information from earlier positions and apply the
recurrence over a longer context.
Masking thus allows us to investigate recurrence
prediction under incomplete observations.
The detailed masking procedure is described in
Appendix~\ref{sec:masking_train}.

These settings provide a controlled way to study when and how Transformers learn recurrence rules. We vary recurrence order, rule organization, input masking, and model size to study generalization, and further examine how the learned rules are represented and executed inside the model.

\subsection{Experiment Setup}
\label{sec:experiment_setup}

We train decoder-only Transformers to predict sequence values
from the observed prefix.
The cross-entropy loss is computed only at designated
prediction positions.
Specifically, for order-$k$ recurrences, the prediction
targets are $q_{k+1},\ldots,q_T$ in the single-rule setting,
$q_{k+2},\ldots,q_T$ in the inter-trajectory setting,
and $q_{k+1},\ldots,q_T$ in the intra-trajectory setting.
Each sequence value is represented by one token.
The vocabulary contains $p$ value tokens,
$\texttt{num\_rule}$ rule-tag tokens,
and, when masking is used, one additional $\bot$ token.

\textbf{Model architecture.}
We use the same decoder-only Transformer architecture in all experiments and vary only the model width $d_{\mathrm{model}}$, number of layers $L$, and number of attention heads $H$. Each Transformer block uses pre-LayerNorm, causal self-attention with rotary position embeddings (RoPE), and a two-layer MLP with GELU activation and hidden dimension $8d_{\mathrm{model}}$. A final LayerNorm is applied before the output layer, whose weights are tied to the input token embeddings. We use no dropout and initialize model parameters from $\mathcal{N}(0,0.02^2)$ unless otherwise specified.

\textbf{Dataset.}
We use $p=127$ unless otherwise specified. For the order-$k$ single-rule setting, there are $p^k$ possible initial states. For the inter-trajectory setting, each initial state can be paired with each rule. We use $70\%$ of these examples for training and the rest for testing. We also ensure that no length-$k$ subsequence in a test sequence appears in any training sequence. Details are given in Appendix~\ref{sec:data_split}. For settings with a much larger data space, including high-order ($k\geq3$) and intra-trajectory settings, we randomly sample $300{,}000$ training examples and use $256$ held-out examples for evaluation. In the intra-trajectory setting, each rule tag is sampled uniformly from the rule pool. When masking is used, we mask only tokens for which the remaining observed prefix still uniquely determines the prediction target, and rule tags are never masked.

\textbf{Training.}
We train all models with cross-entropy loss 
, summed over all prediction positions. Unless otherwise specified, we use AdamW with an initial learning rate of $3\times10^{-4}$, weight decay $1.0$, batch size $512$, and cosine learning-rate decay over $6{,}000$ epochs. We report token-level test accuracy, averaged over all prediction positions.

%% file: capability.tex
\section{Capability Study}
\label{sec:model_training}

Starting from the order-$2$ single-rule setting, we investigate
how Transformer's performance is affected by three task factors:
the number of recurrence rules, the maximum length of consecutive missing/masking observations, and the recurrence order.
Our baseline model has two Transformer layers and
$d_{\mathrm{model}}=1024$. This choice of parameters coincides with a theoretical expressiveness analysis shown in Theorem \ref{thm:theory:two-layer}.
By varying model depth and width, we compare learning curves
across different task configurations and characterize
empirical trends in how Transformer learns these tasks. 
All figures in this section report \emph{test} accuracy.

\subsection{Learning Multiple Recurrence Rules}
\label{sec:multi_rule}

We first study how the number of recurrence rules
affects Transformer learning in the intra-trajectory
and inter-trajectory settings, comparing explicitly
provided rule labels and latent-rule identification. We denote the number of rules by $N$.

\begin{figure}[h]
    \centering
    \includegraphics[width=\linewidth]
    {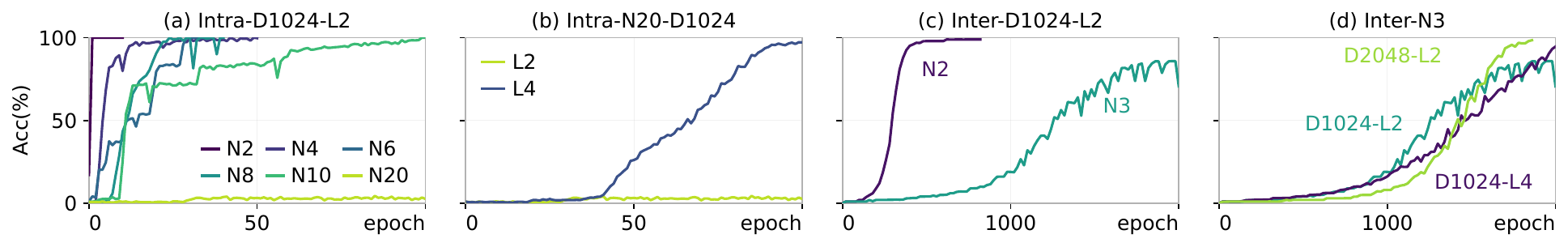}
    \caption{
    Effects of the number of rules and model capacity
    on multi-rule learning.
    Panels (a) and (b) show the intra-trajectory setting:
    (a) compares different $N$ using the baseline
    model size, while
    (b) compares 2 and 4 layers for $N=20$.
    Panels (c) and (d) show the inter-trajectory setting:
    (c) compares $N=2$ and $N=3$ using the baseline
    model, while (d) compares
    three model sizes
     for $N=3$.
    All panels report test accuracy over training epochs. 
    }
    \label{fig:multi_rule}
\end{figure}

\textbf{Intra-trajectory settings are substantially easier to learn.}
Explicit rule labels simplify multi-rule learning in the intra-trajectory
setting, allowing our baseline model to successfully learn a mixture
of 10 rules (Figure~\ref{fig:multi_rule}(a)).
In contrast, in the inter-trajectory setting, where the
model must infer the active rule from context, the same
baseline learns only two rules
(Figure~\ref{fig:multi_rule}(c)).
These results highlight the additional difficulty
introduced by latent rule identification.

Increasing model capacity further improves multi-rule
learning in both settings.
In the intra-trajectory setting, increasing the model depth
from two to four layers enables successful learning
of 20 rules (Figure~\ref{fig:multi_rule}(b)).
In the inter-trajectory setting, increasing either
model depth or width enables successful learning
of three rules (Figure~\ref{fig:multi_rule}(d)).

However, none of the tested model configurations
successfully learns four rules through joint training
in the inter-trajectory setting
(see Appendix~\ref{app:fail_N4}).
Interestingly, this limitation can be overcome through
curriculum learning, which enables the baseline model
to learn four rules by progressively introducing
new rules. We elaborate on this in
Section~\ref{sec:curriculum}.

\subsection{Learning with Masked Observations}
\label{sec:mask_results}

We next investigate the performance under missing
observations. We control masking difficulty through
the maximum permitted length of consecutive masked values,
denoted by $M$, and examine the effects of model depth
$L$ and width $d_{\mathrm{model}}$.

\begin{figure}[h]
    \centering
    \includegraphics[width=\linewidth]
    {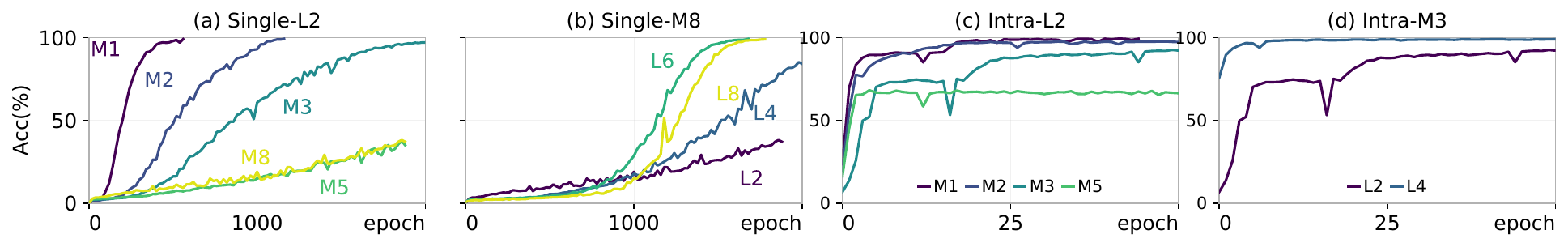}
    \caption{
    Effects of masking difficulty and model depth
    in the single-rule and intra-trajectory settings.
    Panel (a) shows the single-rule setting with
    $L=2$, comparing maximum mask lengths
    $M=1,2,3,5,8$.
    Panel (b) fixes $M=8$ in the single-rule setting
    and compares model depths $L=2,4,6,8$.
    Panel (c) shows the intra-trajectory setting with
    $L=2$, comparing $M=1,2,3,5$.
    Panel (d) fixes $M=3$ in the intra-trajectory
    setting and compares $L=2$ and $L=4$.
    All panels report test accuracy over training epochs.
    }
    \label{fig:mask_depth}
\end{figure}

\textbf{Increasing model depth enables successful generalization under masking.}
In the single-rule setting, larger $M$ increase
learning difficulty (Figure~\ref{fig:mask_depth}(a)). With model width fixed, increasing depth enables
successful learning under masking configurations that
shallower models struggle with
(Figure~\ref{fig:mask_depth}(b)). We observe a similar trend in the intra-trajectory setting
(Figure~\ref{fig:mask_depth}(c)--(d)). These results demonstrate that
increasing depth can improve generalization capability
under incomplete observations across both settings.

\textbf{Width provides an alternative approach.}
Increasing width also enables successful generalization
under masking in the single-rule setting
(Figure~\ref{fig:single_mask_width}).
Our attention analysis (details in Appendix~\ref{app:shortcut_identification})
reveals that models can
retrieve values from alternative historical positions
and employ equivalent recurrence formulas when
immediately preceding observations are masked.
For example, under the rule
$q_t=q_{t-1}+q_{t-2}$,
given the observed pattern
$[\bot,q_6,q_7,\bot]$,
the model can directly predict $q_9$ using
$q_9=2q_7+q_6$,
without recovering the masked values.
By encoding multiple equivalent recurrence formulas in the hidden embedding in parallel (see Theorem \ref{thm:theory:two-layer} for an example),
the model can handle a wider range of masking patterns
without explicitly reconstructing missing observations.
This provides an alternative to increasing depth.

\begin{figure}[h]
    \centering
    \includegraphics[width=0.8\linewidth]
    {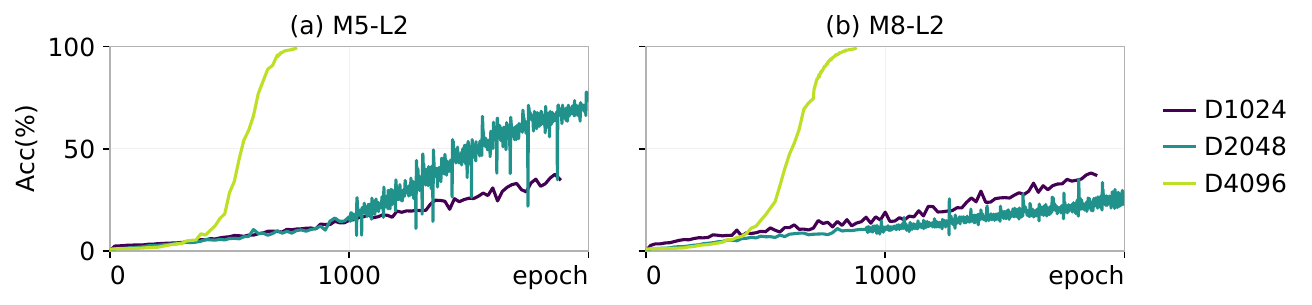}
    \caption{
    Effects of model width on masked prediction
    in the masked single rule setting ($L=2$).
    Panels (a) and (b) show the test accuracies corresponding to $M=5$ and $M=8$, respectively.
    Each compares
    $d_{\mathrm{model}}=1024,2048,4096$.
    }
    \label{fig:single_mask_width}
\end{figure}

Combining masking with the inter-trajectory setting is even harder,
and the model requires additional layers to generalize.
Related experiments and learning curves
are provided in Appendix~\ref{app:fail_mixed_mask}.

\subsection{Learning Higher-Order Recurrences}
\label{sec:higher_order}
We next study how recurrence order affects Transformer learning.
A key feature of higher-order recurrences is that they may admit
simpler equivalent recurrences, which we call \emph{shortcuts}.

Rather than relying on the most recent $k$ values as in the
original order-$k$ recurrence, a shortcut accesses values
further back in history while using fewer than $k$ values
to compute each prediction.
A shortcut using exactly $n$ historical values is called
an \emph{$n$-term shortcut}.

Suppose a shortcut requires access to the previous $k'$
positions, where $k'>k$.
While the original recurrence can make its first prediction
from $q_1,\ldots,q_k$, the shortcut requires
$q_1,\ldots,q_{k'}$ before it can be applied.
Consequently, the shortcut is unavailable for the first
$k'-k$ predictions.
We define this difference as its \emph{delay}
and categorize rules according to their minimum shortcut delay.

For example, the order-$3$ recurrence
$q_t=q_{t-1}+q_{t-2}+q_{t-3}\pmod p$
also admits
\begin{equation}
    q_t=2q_{t-1}-q_{t-4}\pmod p,
    \qquad t\geq5.
    \label{eq:shortcut_example}
\end{equation}
This \emph{two-term shortcut} accesses $q_{t-4}$,
which lies beyond the original three-step context,
but requires only two operands instead of three.
It first applies at $q_5$ rather than $q_4$,
giving a delay of $1$.

While shortcuts reduce the number of historical
values required for prediction, they come at the
cost of accessing more distant positions.
This trade-off suggests that the delay of a
shortcut may play an important role in how
Transformers learn higher-order recurrences.
To investigate this, we categorize order-$3$ recurrence
tasks according to their shortcut delays
and examine the resulting learning behaviors.

\begin{figure}[h]
\centering
\includegraphics[width=\linewidth]{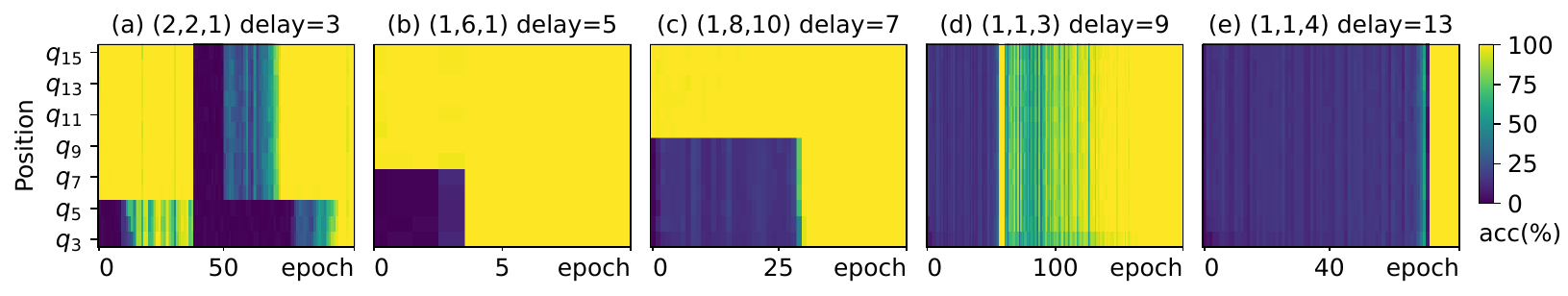}
\caption{
Per-position test accuracy for order-$3$ recurrences
with increasing shortcut delays.
Panels (a)--(e) are arranged by increasing delay
($3,5,7,9,13$), illustrating different learning
behaviors across shortcut delays.
}
\label{fig:shortcut_delay}
\end{figure}

\begin{figure}[h]
\centering
\includegraphics[width=0.9\linewidth]{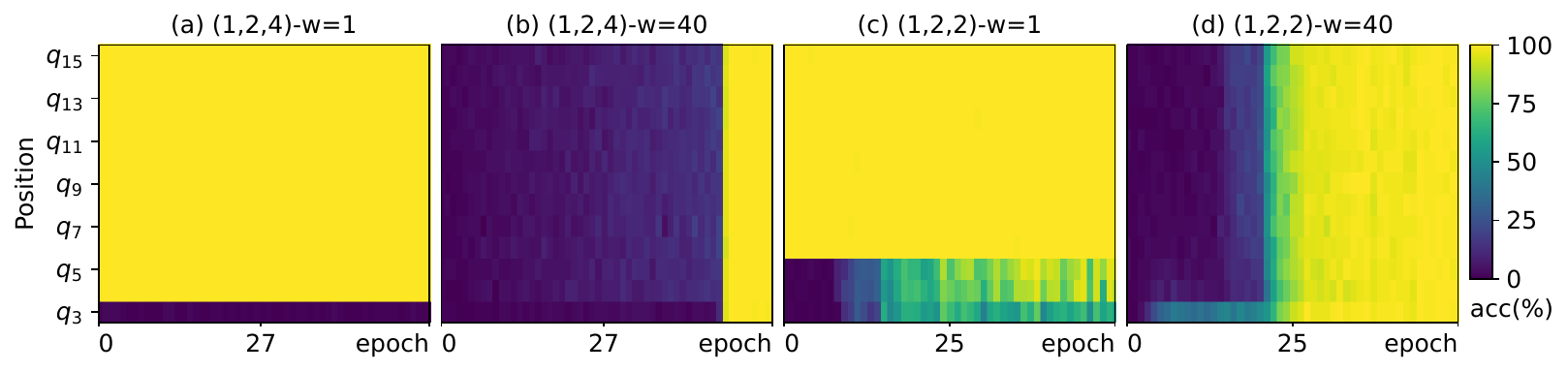}
\caption{Learning dynamics with a sparse shortcut. Each panel shows the test accuracy at each prediction position (y-axis) over training epochs (x-axis). Panels (a)(c) use the original loss weight, while (b)(d) impose 40 times the loss at the first token ($q_4$) prediction. (a)(b) use rule of delay $1$, while (c)(d) use rule of delay $3$.
}
\label{fig:shortcut}
\end{figure}

\textbf{Shortcut delay determines the learned solution.}
We observe three learning patterns as shortcut delay increases.
With small delays, models readily learn shortcuts but struggle
to learn the original recurrence
(Figures~\ref{fig:shortcut}(a)--(c) and
\ref{fig:shortcut_delay}(a)).
With intermediate delays, models first learn shortcuts and later
learn the original recurrence
(Figure~\ref{fig:shortcut_delay}(b)--(c)).
With large delays, models tend to learn the original recurrence
directly, without any shortcut phase
(Figure~\ref{fig:shortcut_delay}(d)--(e)).
Details in how we recognize learned rules are in Appendix~\ref{app:shortcut_identification}.

\textbf{Suppressing shortcuts through loss reweighting.}
Under the standard loss, errors at early prediction positions
can be neglected by the low loss achieved at later positions
using shortcuts. To address this, we increase the loss weight
of the first prediction
(predicting $q_{k+1}$ for order-$k$)
by a factor of 40. As shown in Figure ~\ref{fig:shortcut},
loss reweighting suppresses the shortcut-learning phase
and produces accuracy patterns consistent with learning
the original order-$3$ recurrence across all prediction
positions. However, this intervention comes at a cost:
generalization becomes slower at later positions where
shortcuts would otherwise provide an easier solution.

\textbf{Higher-order tasks remain challenging beyond order-$3$ single-rule setting.}
For order-$4$ tasks, models generalize in our successful cases where two-term shortcuts exist, while the other tested rules are much
harder to learn (Appendix~\ref{app:fail_order_4}).
Combining higher-order recurrences with multiple rules further
increases the difficulty, and the tested intra- and inter-trajectory
models generally fail to generalize
(Appendices~\ref{app:fail_3_intra}--\ref{app:fail_3_inter}).



%% file: mechanism.tex
\section{Curriculum Learning Enhances the Generalization Capability}
\label{sec:curriculum}
The generalization capability studied above raises a natural question: are they fixed by model size and task difficulty, or can they be changed by the training procedure? To study this question, we use curriculum learning, which trains a model on easier tasks before moving to harder ones \citep{bengio2009curriculum}. We consider two ways to increase task difficulty: adding rules one by one and adding input masking after learning the unmasked task. In each stage, we move to the next task once the test accuracy on the current task reaches a certain threshold.

\begin{figure}[htbp!]
\centering
\includegraphics[width=0.7\linewidth]{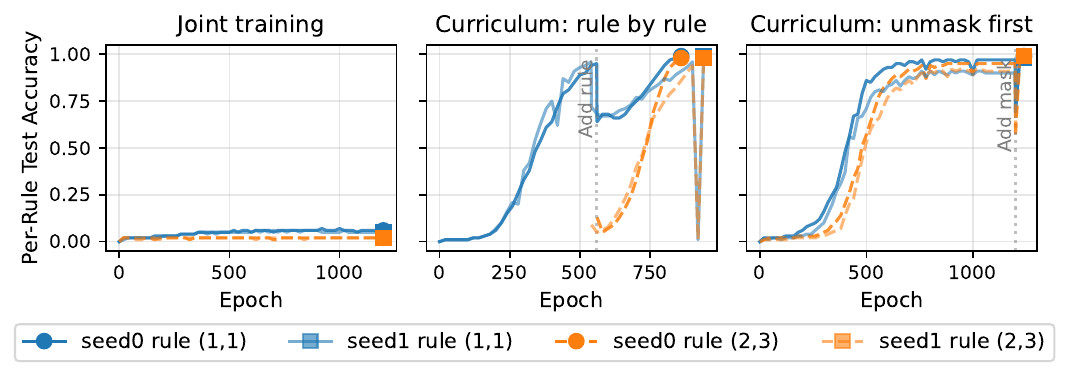}
\caption{Joint training and curriculum training on the masked inter-trajectory two-rule task with $10\%$ random input masking. We compare joint training (\textit{left}) with two curricula: adding rules one by one under masking (\textit{middle}), and learning both rules without masking before adding random masking (\textit{right}).}
\label{fig:two_task_figure}
\end{figure}

\textbf{Curriculum learning enables generalization when joint training fails.}
We first study the masked inter-trajectory two-rule task, where joint training fails to generalize to both rules within $1{,}200$ epochs. As shown in Figure~\ref{fig:two_task_figure}, both curricula reach over $98\%$ test accuracy ($d_{\mathrm{model}}=1024$, $H=4$, $L=4$). In the rule-by-rule curriculum, adding the second rule ($a^{(2)}=(2,3)$) first reduces accuracy on the first rule ($a^{(1)}=(1,1)$), but continued training recovers its accuracy while learning the second rule. In the masking curriculum, switching from unmasked to masked inputs also causes an accuracy drop, but the model recovers to $98\%$ within $40$ epochs.

\begin{figure}[h]
    \centering
    \includegraphics[width=0.8\linewidth]{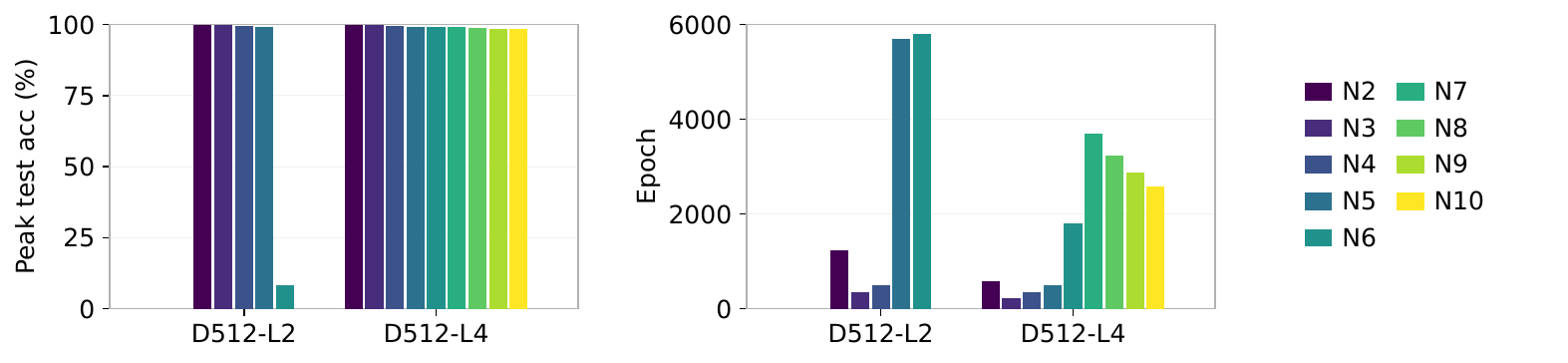}
    \caption{
    Curriculum learning in the inter-trajectory setting. The number of rules ($N$) increases progressively from 2 to 10. The left panel reports peak test accuracy, and the right panel shows the corresponding minimum training epochs reaching $98\%$ test accuracy for two- and four-layer models.
    }
    \label{fig:curriculum}
\end{figure}

\textbf{Curriculum learning extends generalization to more rules.}
We next apply the rule-by-rule curriculum to inter-trajectory tasks with more rules. We start with two rules and add one new rule whenever the model reaches $95\%$ test accuracy, while keeping all previously introduced rules in training. Under joint training, models generalize to at most $4$ rules, even with $L=6$ or $d_{\text{model}}=4096$. In contrast, curriculum learning enhances this capability to $5$ rules with $L=2$ and $10$ rules with $L=4$, as shown in the left panel of Figure~\ref{fig:curriculum}. 

\textbf{Deeper models require fewer epochs under the curriculum.}
The benefit of model depth is also reflected in training speed. As shown in the right panel of Figure~\ref{fig:curriculum}, models with $L=4$ require fewer epochs than models with $L=2$ to learn the same number of rules. Thus, within the model sizes studied here, increasing depth both extends the number of rules that can be learned and reduces the training epochs required under curriculum learning.

\section{How Transformers Learn, Represent, and Execute Rules}
\label{sec:rule_mechanisms}
The previous sections study when Transformers generalize under different model sizes and training procedures. We now use smaller settings where models reliably generalize to examine data interactions between rules, depth-dependent rule representations, and whether models follow the intended local recurrence.

\subsection{Data Interactions Between Two Rules}
\label{sec:rule_data_interactions}

We study how the amount of training data for one rule affects generalization on another in the inter-trajectory two-rule setting. For each rule $i$, we select a fraction $r_i$ of its $p^2$ available examples for training and vary the two inter-trajectories independently to examine their effects on per-rule held-out test accuracy. Figure~\ref{fig:rule_data_interactions} reports results for order-$2$ rules. Details are provided in Appendix~\ref{app:rule_data_interactions}.

\textbf{Data from another rule can either hinder or help generalization.}
Figure~\ref{fig:rule_data_interactions} shows that the effect of adding data from another rule is non-monotonic. (i) \emph{Negative transfer:} adding a small amount of data from another rule can prevent generalization. For example, the second rule generalizes at $(r_1,r_2)=(0,0.3)$ but fails at $(0.05,0.3)$. (ii) \emph{Positive transfer:} sufficient data from another rule can enable generalization with otherwise insufficient data. For example, the first rule fails at $(r_1,r_2)=(0.3,0)$ but generalizes at $(0.3,0.5)$. Thus, data from different rules can either compete with or support each other during learning.

\begin{figure}[htbp!]
\centering
\includegraphics[width=0.9\linewidth]{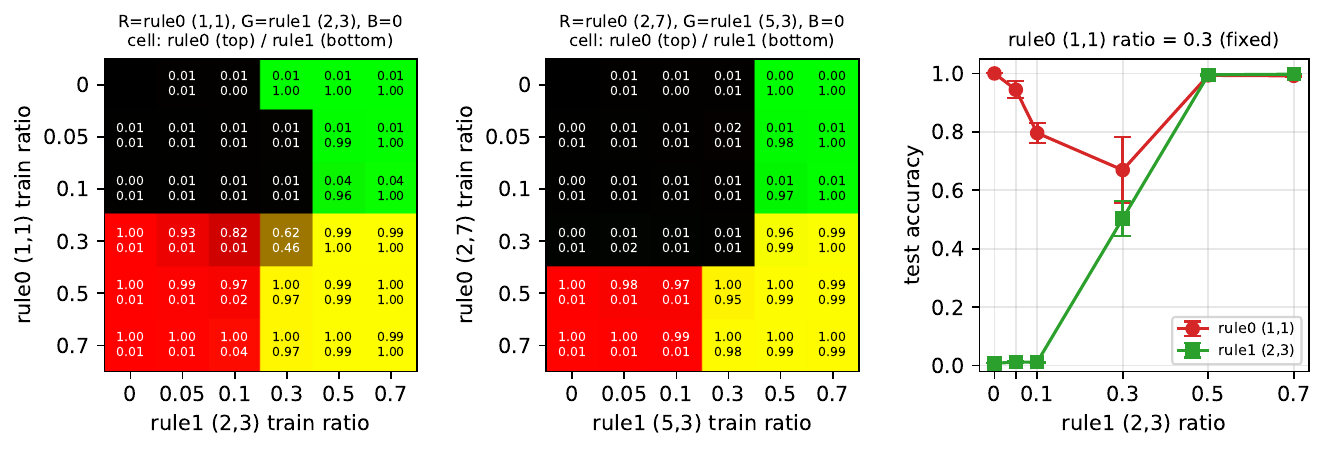}
\caption{Data from another rule can hinder or facilitate generalization. (\textit{Left, middle}) Per-rule held-out test accuracy as both training fractions vary, including single-rule controls at zero training fraction. (\textit{Right}) Non-monotonic generalization as one rule's training fraction increases while the other's is fixed; curves show mean $\pm$ standard deviation over four seeds.}
\label{fig:rule_data_interactions}
\end{figure}

\subsection{Depth and Rule Representations}
\label{sec:depth_rule_representations}
We next examine how model depth affects both rule learning and internal representations.

\textbf{Rule identity becomes linearly separable with depth.}
We use linear probes to test whether the rule can be decoded from internal representations in the unmasked two-rule inter-trajectory setting. As shown in Table~\ref{tab:rule_probes_all_layers}, probes on single-layer models and the first layer of two-layer models remain near chance, while the second layer of two-layer models makes rule identity highly linearly decodable, especially from MLP hidden activations. This suggests that linearly separable rule representations emerge with depth. Probe details are provided in Appendix~\ref{app:rule_probes}.

\begin{figure}[h]
\centering
\includegraphics[width=0.8\linewidth]{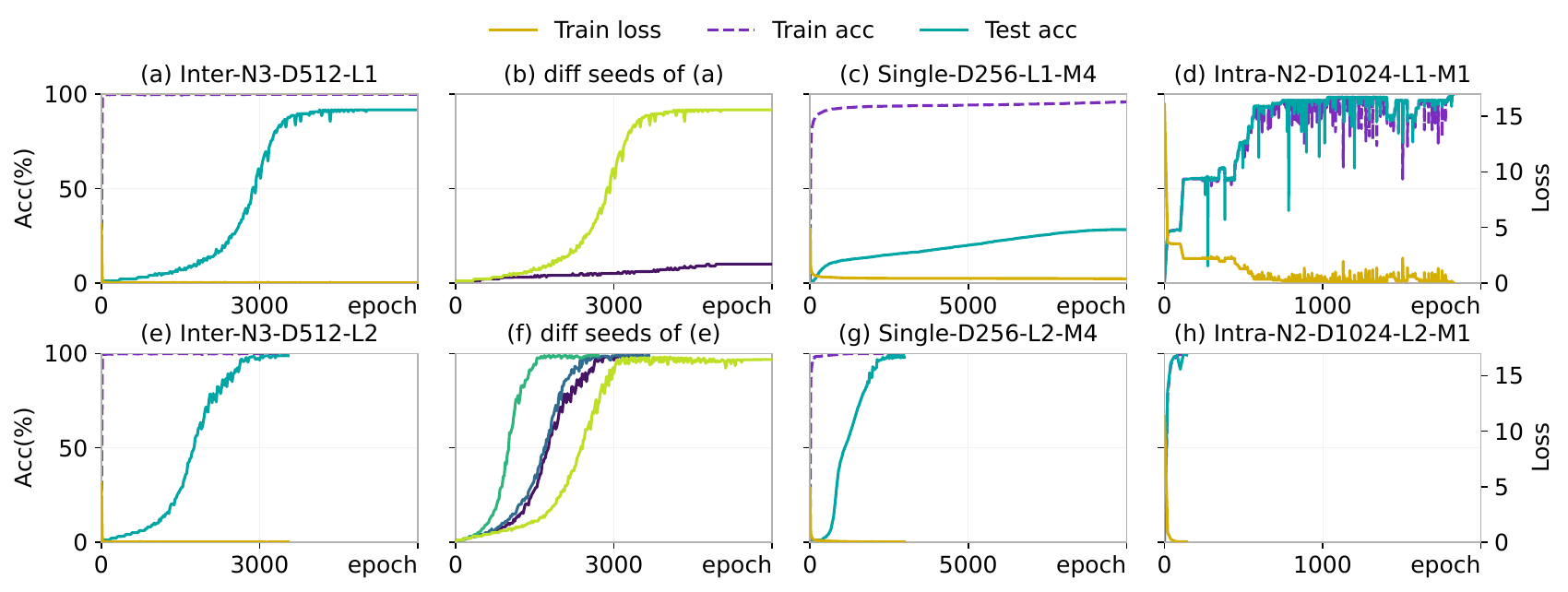}
\caption{Single-layer Transformers take longer to generalize than two-layer Transformers across task settings. The top and bottom rows show one- and two-layer models, respectively. The first two columns show the three-rule inter-trajectory setting, the third shows the single-rule setting with length-$4$ masking, and the last shows the two-rule intra-trajectory setting with length-$1$ masking. In (b) and (f), the additional solid curves show test accuracy across four training seeds.}
\label{fig:L1L2cmp}
\end{figure}

\textbf{Single-layer models take substantially longer to generalize.}
Model depth also affects the speed of generalization. Across different task settings, single-layer Transformers require substantially more training epochs to generalize than two-layer Transformers (Figure~\ref{fig:L1L2cmp}). This gap is consistent across inter-trajectory, single-rule masked, and intra-trajectory masked settings.

\subsection{Do Models Follow the Intended Recurrence?}
\label{sec:local_recurrence}
High test accuracy does not reveal how a model computes its predictions. Since each valid sequence is deterministic, a model could predict from the early trajectory rather than apply the recurrence to recent values. We therefore test the \emph{local recurrence} hypothesis by breaking the early trajectory while keeping the information needed for local prediction. In the inter-trajectory setting, this information must also identify the rule, while in the intra-trajectory setting the rule is given by $c_t$. Details are provided in Appendix~\ref{app:random_value_recurrence}.

\textbf{Inter-trajectory models follow the local recurrence.}
For the unmasked order-$2$ two-rule setting, we construct inputs with a \textit{random} early prefix followed by a short suffix generated by one rule. Across five checkpoints with one or two layers, prediction accuracy is at least $98\%$ for each rule (Table~\ref{tab:inter-trajectory_local_recurrence}). Thus, the models can continue the recent recurrence even when the earlier prefix does not follow the same rule.


\textbf{Unmasked intra-trajectory models follow the local recurrence.} We independently randomize all value tokens ($q_t$) while keeping the rule tags ($c_t$) and test whether the model predicts $f_{\sigma_t}(q_{t-2},q_{t-1})$ at each rule tag $c_t$. All 4 unmasked runs reach $100\%$ agreement with the local recurrence (Table~\ref{tab:intra_trajectory_local_recurrence}), showing that the models directly apply the given rule to the two recent values. 

\textbf{Masking makes rule execution less local.} For the order-2 two-rule intra-trajectory setting, with $10\%$ length-$1$ input masking during training, half of the runs no longer consistently follow the local recurrence, with the failures occurring only for rule $(1,1)$ (Table~\ref{tab:intra_trajectory_local_recurrence}). Since all values are visible during evaluation, this suggests that some masked models use information beyond the two most recent values.

Attention scores alone do not show which tokens are needed for prediction. The attention maps in Figures~\ref{fig:inter-trajectory-attention}--\ref{fig:intra-trajectory-attention-mask} show nonzero attention to earlier tokens, while the tests above show that a valid early trajectory is not required for local prediction. We therefore use the intervention results, rather than attention scores alone, as the main evidence for local recurrence.

%% file: conclusion.tex
\section{Conclusion}
\label{sec:conclusion}

We study when and how Transformers learn multiple recurrence rules in a controlled setting. Our results lead to three main observations. First, the generalization capability is not determined by model size alone: rule organization, algebraic structure, masking, and training procedure can all change which tasks are learned. Second, successful prediction does not imply that the model learns the intended computation, as models may instead use algebraic shortcuts or other equivalent recurrence formulas. Third, depth affects both learning and representation: two-layer models learn faster and develop linearly decodable rule representations that are not observed at the probed sites in single-layer models.

These findings open several directions for further study. In particular, understanding how single-layer models perform multi-rule prediction may help explain why linearly separable rule representations emerge with depth. Extending these questions to larger models and more complex sequence tasks may further reveal how Transformers learn and organize multiple rules.

%% file: AI_use_statement.tex
\subsection*{AI use statement}

In this work, we used generative AI tools to generate
synthetic datasets,  implement
methods, assist with translation, clean and reformat
datasets, design or provide feedback
on research methodology or experiments, assist in writing proofs,
formulate mathematical
claims, provide critical ingredients for proving mathematical
claims, and support qualitative and thematic data analysis.

We did not use generative AI tools to develop theoretical
models or conceptual frameworks, propose or refine hypotheses, or interpret results.

Additionally, we used generative AI tools to create and
modify scientific figures, edit software code, summarize
existing literature, brainstorm ideas, format references, 
edit the manuscript to improve readability,
and suggest improvements to its structure..

We have reviewed all AI-assisted work. AI-generated code
was verified and tested for correctness by all authors.
AI-assisted data analysis was independently verified,
and AI-generated plots were checked against the original
experimental data.

We take responsibility for the final content of this work,
including text, claims, and artifacts produced with
the assistance of generative AI.

%% file: appendix.tex
\input{appendix_related_work}

\input{experiment_details}

\input{shortcut_details}

\input{shortcut_verify}

\input{failed_tasks}

\section{Extension to Inter-Trajectory Learning with Nonlinear Recurrence Rules}
\label{sec:non-linear}

We extend the inter-trajectory setting from linear recurrence rules to mixtures of linear and nonlinear rules. We use the addition rule
$$
f_{\mathrm{add}}:\quad
q_t=q_{t-1}+q_{t-2}\pmod{127},
$$

and consider two multiplication rules:

$$
f_{\mathrm{mul}}:\quad
q_t=q_{t-1}q_{t-2}\pmod{127},
\qquad
f_{\mathrm{mul23}}:\quad
q_t=q_{t-1}^2q_{t-2}^3\pmod{127}.
$$

We study two two-rule settings: $(f_{\mathrm{add}},f_{\mathrm{mul}})$ and $(f_{\mathrm{add}},f_{\mathrm{mul23}})$. The second setting follows the same coefficient pattern as the two linear rules $(1,1)$ and $(2,3)$ used in our main experiments, allowing us to compare linear and nonlinear rule mixtures under a similar setup. All models are jointly trained from random initialization without curriculum learning, with either $0\%$ or $10\%$ random input masking.

\paragraph{Data, models, and training.}
We use $70\%$ of the examples from each rule for training and the rest for testing. We evaluate five model configurations:
$(d_{\mathrm{model}},L,H)\in{(256,1,2),(256,2,2),(256,2,4),(512,1,4),(512,2,4)}$.
We use AdamW with an initial learning rate of $3\times10^{-4}$, weight decay $1.0$, batch size $512$, and cosine learning-rate decay over a maximum of $6{,}000$ epochs. Training stops $200$ epochs after the overall test accuracy first reaches $99\%$.

\begin{figure}[htbp!]
\centering
\includegraphics[width=\linewidth]{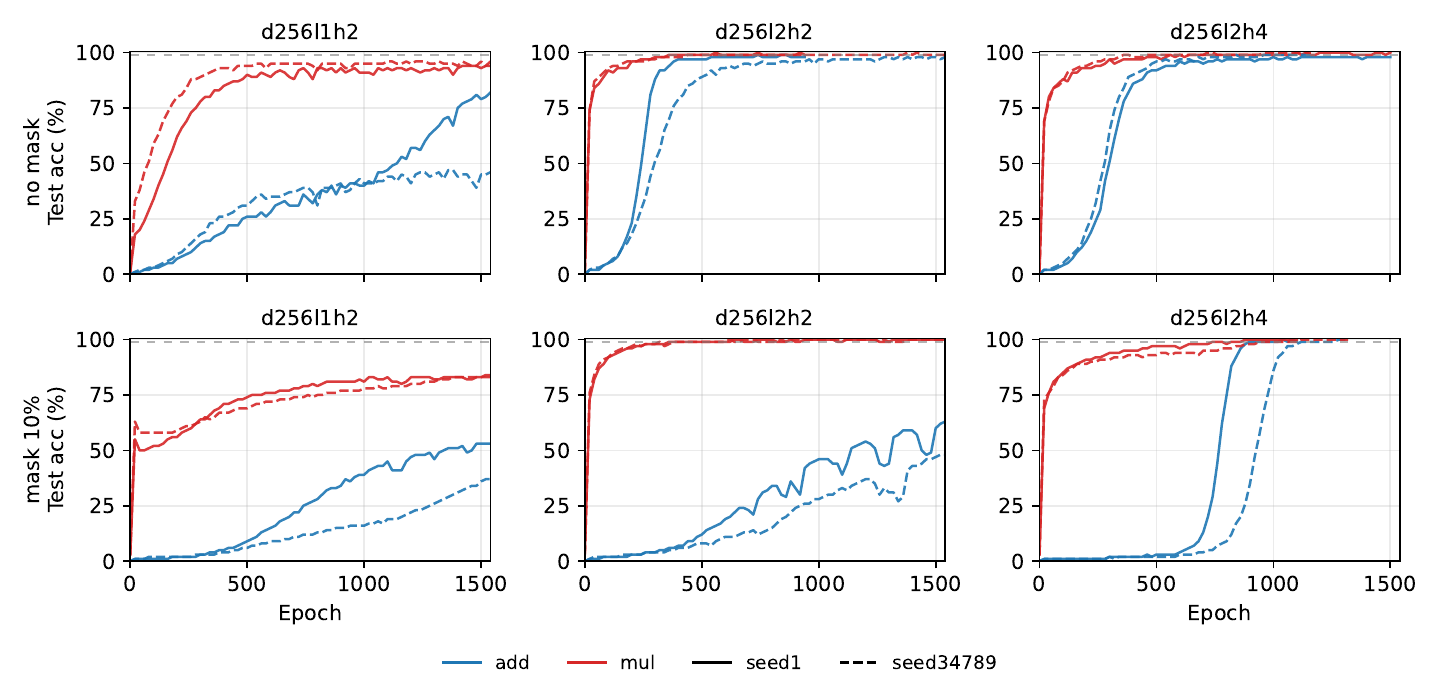}
\caption{Learning curves for $f_{\mathrm{add}}$ and $f_{\mathrm{mul}}$ with $d_{\mathrm{model}}=256$. Subplot titles report $d_{\mathrm{model}}$, $L$, and $H$ after \texttt{d}, \texttt{l}, and \texttt{h}, respectively.}
\label{fig:mul_add_d256}
\end{figure}

\begin{figure}[htbp!]
\centering
\includegraphics[width=0.7\linewidth]{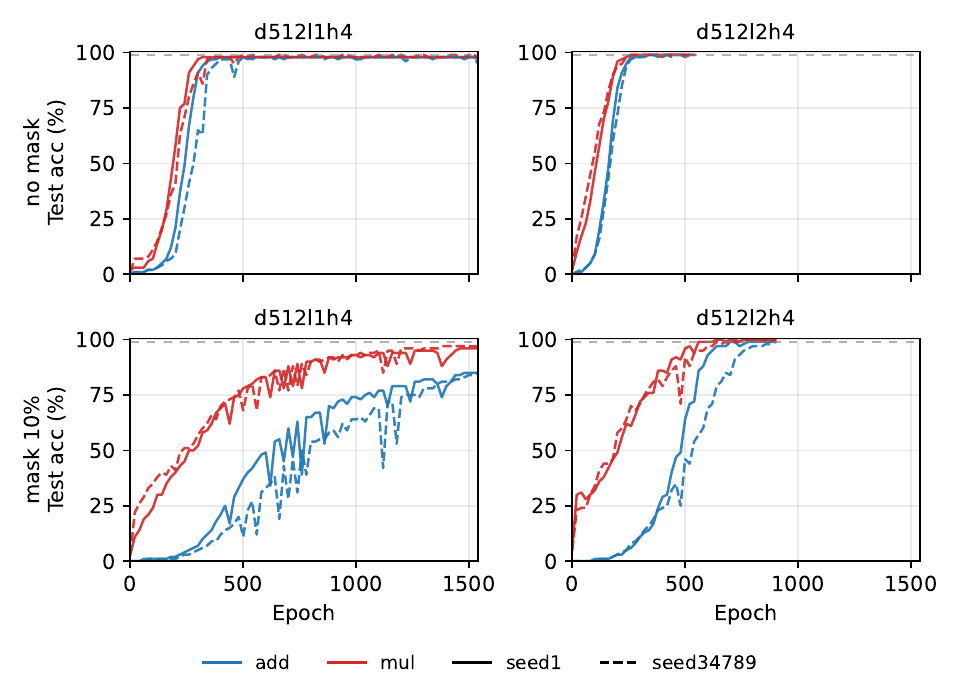}
\caption{Learning curves for $f_{\mathrm{add}}$ and $f_{\mathrm{mul}}$ with $d_{\mathrm{model}}=512$. Subplot titles report $d_{\mathrm{model}}$, $L$, and $H$ after \texttt{d}, \texttt{l}, and \texttt{h}, respectively.}
\label{fig:mul_add_d512}
\end{figure}

\paragraph{Multiplication is learned faster than addition on $(f_{\text{add}},f_{\text{mul}})$.}
Figures~\ref{fig:mul_add_d256} and~\ref{fig:mul_add_d512} show the learning curves for $(f_{\mathrm{add}},f_{\mathrm{mul}})$. Across the evaluated model settings and training runs, multiplication reaches high test accuracy earlier than addition, or has higher accuracy when generalization is incomplete. This pattern appears both with and without input masking, showing that the nonlinear multiplication rule is not harder to learn than addition in this setting.

\begin{figure}[htbp!]
\centering
\includegraphics[width=\linewidth]{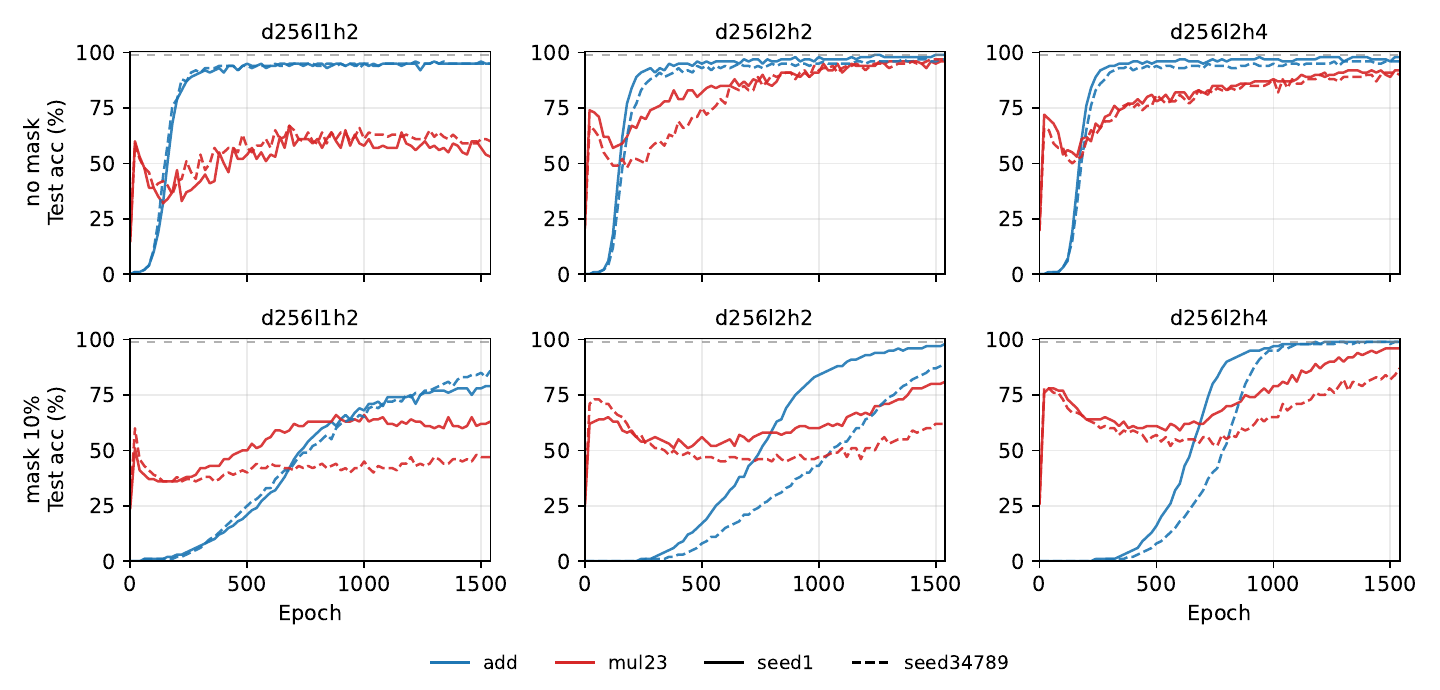}
\caption{Learning curves for $f_{\mathrm{add}}$ and $f_{\mathrm{mul23}}$ with $d_{\mathrm{model}}=256$. Subplot titles report $d_{\mathrm{model}}$, $L$, and $H$ after \texttt{d}, \texttt{l}, and \texttt{h}, respectively.}
\label{fig:mul23_add_d256}
\end{figure}

\begin{figure}[htbp!]
\centering
\includegraphics[width=0.7\linewidth]{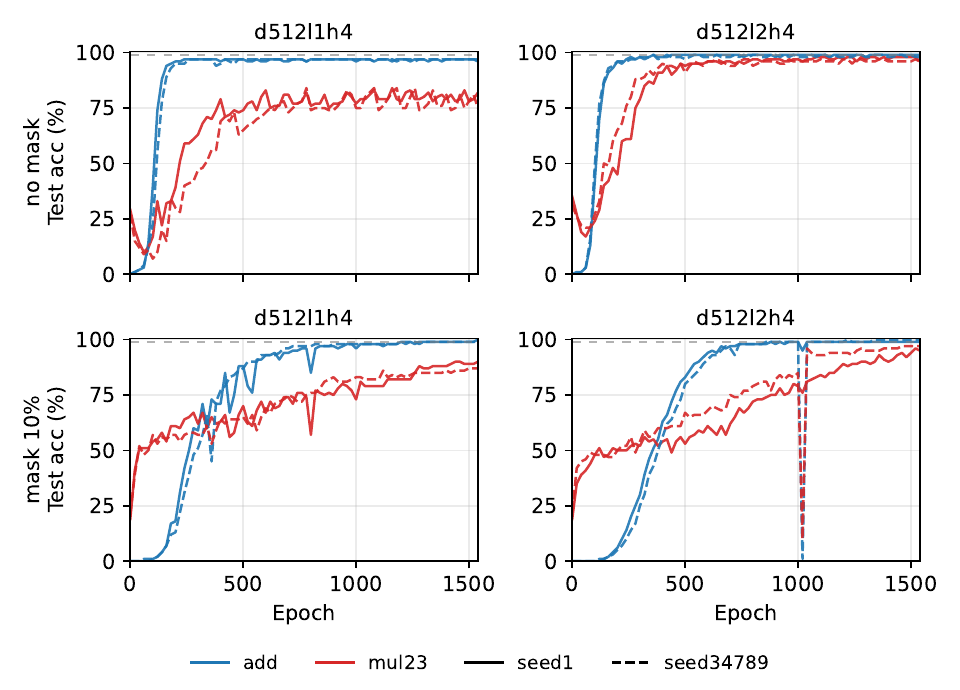}
\caption{Learning curves for $f_{\mathrm{add}}$ and $f_{\mathrm{mul23}}$ with $d_{\mathrm{model}}=512$. Subplot titles report $d_{\mathrm{model}}$, $L$, and $H$ after \texttt{d}, \texttt{l}, and \texttt{h}, respectively.}
\label{fig:mul23_add_d512}
\end{figure}

\paragraph{Input masking delays generalization.}
Generalization occurs later with $10\%$ input masking than without masking, especially for the addition rule. This pattern appears for both nonlinear rule mixtures, as shown by comparing the masked and unmasked results in Figures~\ref{fig:mul_add_d256}--\ref{fig:mul23_add_d512}. This is consistent with our results for linear rules, where input masking also makes generalization harder.

\paragraph{Nonlinear rules do not necessarily make generalization harder.}
Figures~\ref{fig:mul23_add_d256} and~\ref{fig:mul23_add_d512} show that two-layer models can generalize to both $f_{\mathrm{add}}$ and $f_{\mathrm{mul23}}$ even with $10\%$ random input masking. In contrast, joint training on the two linear rules $(1,1)$ and $(2,3)$ does not reach joint generalization under the same masking rate in our main experiments. Thus, the difficulty of inter-trajectory learning cannot be explained by whether the recurrence rules are linear or nonlinear alone.

\section{Experimental details for how transformers learn, represent, and execute rules}

\subsection{Experimental Details for Interactions Between Training Fractions}
\label{app:rule_data_interactions}

\paragraph{Task and data construction.}
The experiments in Figure~\ref{fig:rule_data_interactions} use the inter-trajectory setting with two order-$2$ rules, with coefficient pairs $(1,1)/(2,3)$ and $(2,7)/(5,3)$. For each rule, $10\%$ of the data examples form the held-out test block. Each rule's training fraction is selected from $\{0.05,0.1,0.3,0.5,0.7\}$ and is measured relative to the full $p^2$ examples per rule. The same held-out test dataset is used across fraction combinations for a given seed. 

\paragraph{Model and training.}
We use the decoder-only Transformer described in the main text, with $d_{\text{model}}=512$, $L=2$ layers, and $H=4$ attention heads. Training uses AdamW with learning rate $3\times10^{-4}$, weight decay $1.0$, batch size 512, and a cosine-annealing schedule over $20{,}000$ epochs. Early stopping occurs only after overall held-out test accuracy reaches $99\%$ for 200 consecutive epochs. Per-rule held-out accuracies are reported at the best epoch of each run.

\paragraph{Random seeds.}
The left and middle panels of Figure~\ref{fig:rule_data_interactions} use seed 17996. The right panel reports the mean and standard deviation across seeds $\{17996,18318,34789,44536\}$, with the $(1,1)$ training fraction fixed at $0.3$ and the $(2,3)$ training fraction varied. The rule labels 0 and 1 in the plots correspond to the first and second rules, respectively, whose training fractions are denoted by $r_1$ and $r_2$ in the text.

\subsection{Linear probes of rule identity}
\label{app:rule_probes}

\paragraph{Task and checkpoints.}
The probes in Table~\ref{tab:rule_probes_all_layers} distinguish the two rules in the unmasked inter-trajectory setting. We evaluate six trained models with $(d_{\mathrm{model}},L,H)\in{(256,1,2),(512,1,4),(256,2,2),(256,2,4),(512,2,2),(512,2,4)}$ and sequence length $16$. Each model is jointly trained on the two rules using $70\%$ of the data from each rule. We use the final checkpoint of each run and keep the model frozen during probe training.

\paragraph{Probe data and split.}
For each rule, we generate length-16 sequence windows from all $127^2$ initial states, shuffle them with seed 0, and retain the first 2,000. Within each rule, a further random permutation with seed 0 assigns 1,000 sequences to probe training and 1,000 to probe testing. The two rules are permuted separately. Both splits are therefore balanced, containing 2,000 sequences each; binary labels 0 and 1 encode rules $(1,1)$ and $(2,3)$, respectively. The same split is used across layers and token positions. The held-out designation refers to probe fitting: these examples are drawn from the full state space, without requiring exclusion from the Transformer's original training set or enforcing disjoint recurrence trajectories between the probe splits.

\paragraph{Activation sites and positions.}
At each layer, the attention feature concatenates all $H$ head outputs after attention-weighted value aggregation, immediately before the attention output projection. Its dimension is $d_{\text{model}}$. The MLP feature is the hidden vector after the first linear map and GELU, immediately before the second linear map, with dimension $r d_{\text{model}}$. Both features are computed downstream of the respective sublayer's input LayerNorm and upstream of its output projection and residual addition. Neither is a normalized residual-stream output. In particular, the MLP feature is not the final MLP output. We extract activations in evaluation mode and fit a separate probe for every checkpoint, layer, site, and input position. Each probe receives one vector per sequence, with no concatenation or averaging across positions. We number layers from 1 in the paper. The stored activation keys number them from 0.

\paragraph{Probe fitting and evaluation.}
We standardize each feature coordinate using the probe training set only and apply the fitted transformation to the test set. A logistic-regression classifier with an intercept and $\ell_2$ regularization is then fitted using \texttt{scikit-learn}, with inverse regularization strength $C=1$, solver \texttt{lbfgs}, and at most 1,000 iterations. AUROC uses the predicted probability of rule $(2,3)$. Classification accuracy uses a probability threshold of $0.5$. Chance performance is $0.5$ for both metrics on these balanced splits. The retained results use one probe split per checkpoint, with no repeated-seed uncertainty estimate. The separate five-fold cross-validation notebook in the code archive uses a different sampling protocol and is not the source of the reported numbers.

\paragraph{Aggregation and layer dependence.}
The position-specific results cover all 16 input tokens. Tables~\ref{tab:rule_probes_all_layers} average metrics over $\left\{q_3,\cdots,q_{16}\right\}$, weighting each position equally. This excludes the two initial values, before any recurrence transition has been observed, and includes every subsequent position rather than selecting the best-performing position. Metrics are computed separately at each position before averaging. Predictions from different probes are not pooled to compute a single AUROC. These positional averages are not averages over independent training runs.

Table~\ref{tab:rule_probes_all_layers} reports both metrics at every layer. First-layer AUROC remains near $0.5$ in all six configurations, whereas second-layer MLP AUROC is high in every two-layer model. Attention decodability is more position-dependent: for $(d_{\text{model}},L,H)=(256,2,4)$, second-layer attention AUROC increases from $0.789$ at the third token to $0.933$ at the fourth and at least $0.974$ thereafter; for $(512,2,2)$, it rises from $0.657$ at the third token to $0.993$ at the sixteenth. Thus, averaging across positions includes the early prefixes where rule information is less readily decoded.

\begin{table}[htbp!]
\centering
\small
\caption{Rule-probe results at every layer, averaged over the $\left\{q_3,\cdots,q_{16}\right\}$ input tokens. Attention and MLP hidden features use the sites specified above. Accuracies are fractions. }
\label{tab:rule_probes_all_layers}
\begin{tabular}{cccccc}
\hline
$(d_{\text{model}},L,H)$ & Layer & \multicolumn{2}{c}{AUROC} & \multicolumn{2}{c}{Accuracy} \\
 & & Attention & MLP hidden & Attention & MLP hidden \\
\hline
$(256,1,2)$ & $1$ & $0.492$ & $0.480$ & $0.494$ & $0.487$ \\
$(512,1,4)$ & $1$ & $0.487$ & $0.487$ & $0.490$ & $0.493$ \\
\hline
$(256,2,2)$ & $1$ & $0.491$ & $0.478$ & $0.492$ & $0.484$ \\
 & $2$ & $0.502$ & $0.998$ & $0.502$ & $0.984$ \\
$(256,2,4)$ & $1$ & $0.489$ & $0.489$ & $0.492$ & $0.492$ \\
 & $2$ & $0.967$ & $0.999$ & $0.916$ & $0.990$ \\
$(512,2,2)$ & $1$ & $0.491$ & $0.489$ & $0.495$ & $0.493$ \\
 & $2$ & $0.942$ & $0.997$ & $0.890$ & $0.978$ \\
$(512,2,4)$ & $1$ & $0.492$ & $0.491$ & $0.495$ & $0.494$ \\
 & $2$ & $0.991$ & $0.996$ & $0.978$ & $0.985$ \\
\hline
\end{tabular}
\end{table}

\subsection{Local Recurrence Tests}
\label{app:random_value_recurrence}

\begin{figure}[htbp!]
\centering
\includegraphics[width=0.8\linewidth]{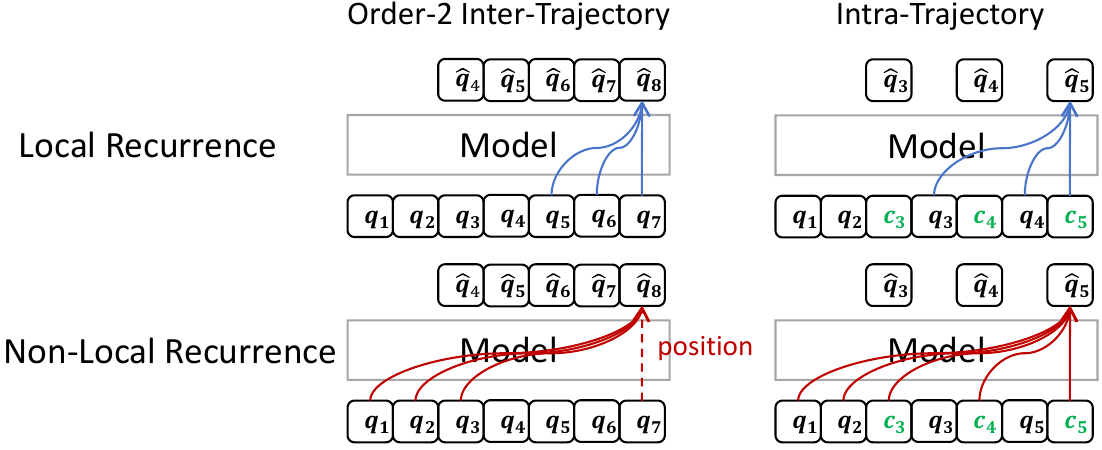}
\caption{Illustration of the local recurrence test. We break the earlier trajectory while keeping the recent values needed for local prediction. In the inter-trajectory setting, the recent values must also identify the rule. In the intra-trajectory setting, the rule is given by the current rule tag.}
\label{fig:local-recurrence-illustration}
\end{figure}

We test whether the models use the intended local recurrence by breaking the earlier trajectory while keeping the information needed for local prediction. Figure~\ref{fig:local-recurrence-illustration} illustrates the test. The construction differs slightly between the inter- and intra-trajectory settings because the rule must be inferred from values in the former but is directly given by the rule tag in the latter.

\subsubsection{Two-rule inter-trajectory setting}

We construct inputs with a random early prefix followed by a short rule-consistent suffix. Using sampling seed $42$, we sample $200$ prefixes with $q_1,\ldots,q_6$ independently and uniformly drawn from $\mathbb{F}_{127}$. For each rule $f_a$, we then generate
$$
q_7=f_a(q_5,q_6),
\qquad
q_8=f_a(q_6,q_7),
$$
and test whether the model predicts
$$
q_9=f_a(q_7,q_8).
$$

The model receives $(q_1,\ldots,q_8)$ without a rule tag. Thus, the full trajectory is generally invalid, while the recent values $(q_5,q_6,q_7,q_8)$ follow the selected rule. Table~\ref{tab:inter-trajectory_local_recurrence} reports the prediction accuracy. We use the same random prefixes for all rules and checkpoints.

\begin{table}[htbp!]
\centering
\caption{Accuracy on the local recurrence test in the two-rule inter-trajectory setting. Each model is evaluated on $200$ random prefixes per rule.}
\label{tab:inter-trajectory_local_recurrence}
\begin{tabular}{ccccc}
\hline
$d_{\text{model}}$ & $L$ & $H$ & Rule $(1,1)$ & Rule $(2,3)$ \\
\hline
$256$ & $2$ & $2$ & $0.995$ & $0.995$ \\
$256$ & $2$ & $4$ & $0.995$ & $0.995$ \\
$512$ & $1$ & $4$ & $0.995$ & $1.000$ \\
$512$ & $2$ & $2$ & $0.995$ & $1.000$ \\
$512$ & $2$ & $4$ & $0.980$ & $0.985$ \\
\hline
\end{tabular}
\end{table}

We further test whether the prediction depends on the value before the recent rule-consistent triple. We vary $q_5$ over all $127$ values while keeping all other input values, including $(q_6,q_7,q_8)$, fixed. This breaks the transition into $q_7$ for most values of $q_5$ while preserving the final triple $(q_6,q_7,q_8)$ and the target $q_9=f_a(q_7,q_8)$. Across the $25{,}400$ inputs per rule and checkpoint, accuracy ranges from $91.6\%$ to $99.6\%$; all models except the $d_{\mathrm{model}}=512$, $L=2$, $H=4$ model remain above $95\%$. Thus, the recent valid triple is enough for high prediction accuracy, although predictions are not fully independent of the earlier context.

\subsubsection{Two-rule intra-trajectory setting}

We test local recurrence more directly in the intra-trajectory setting by randomizing all value tokens. The input keeps the original form
$$
[q_1,q_2,c_3,q_3,c_4,q_4,\ldots],
$$
where each $q_t$ is sampled independently and uniformly from $\mathbb{F}_{127}$ and each rule tag $c_t$ is sampled as in training. Let $\sigma_t$ denote the rule given by $c_t$. No value is masked during evaluation, including for models trained with input masking.

At each rule tag $c_t$, we compare the model prediction $\hat q_t$ with the local target
$$
q_t^{\mathrm{local}}
=
f_{\hat{c}_t}(q_{t-2},q_{t-1}).
$$
Because the value tokens are sampled independently, this target generally differs from the random token $q_t$ that follows $c_t$. For $B$ test sequences, we measure

$$
\mathrm{Acc}_{\mathrm{local}}(t)
=
\frac{1}{B}\sum_{b=1}^{B}
\mathbf{1}\!\left[
\hat q_t^{(b)}
=
f_{\hat{c}_t^{(b)}}\!\left(
q_{t-2}^{(b)},q_{t-1}^{(b)}
\right)
\right].
$$

Predictions are evaluated from the supplied prefix and are not fed back into later positions.

\begin{table}[htbp!]
\centering
\caption{Agreement with the local recurrence on randomized-value inputs in the two-rule intra-trajectory setting. Each row represents one training run.}
\label{tab:intra_trajectory_local_recurrence}
\begin{tabular}{ccccc}
\hline
Mask rate & Run & All rules & Rule $(1,1)$ & Rule $(2,3)$ \\
\hline
$0\%$ & A & $1.00$ & $1.00$ & $1.00$ \\
$0\%$ & B & $1.00$ & $1.00$ & $1.00$ \\
$0\%$ & C & $1.00$ & $1.00$ & $1.00$ \\
$0\%$ & D & $1.00$ & $1.00$ & $1.00$ \\
\hline
$10\%$ & A & $0.64$ & $0.29$ & $1.00$ \\
$10\%$ & B & $1.00$ & $1.00$ & $1.00$ \\
$10\%$ & C & $0.80$ & $0.59$ & $1.00$ \\
$10\%$ & D & $1.00$ & $1.00$ & $1.00$ \\
\hline
\end{tabular}
\end{table}

\subsubsection{Visualization of Attention Scores}
\label{app:attention_visualization}

\begin{figure}[htbp!]
\centering
\includegraphics[width=0.95\linewidth]{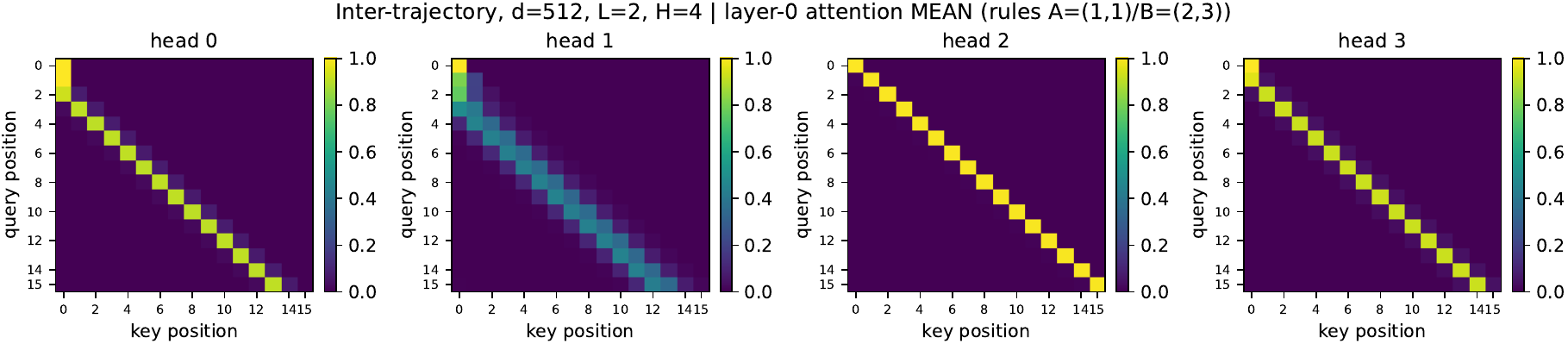}
\caption{Attention scores in the unmasked two-rule inter-trajectory setting. Earlier values also receive nonzero attention; for example, Head 1 attends to $q_{t-4}$ when predicting $q_t$ at position $q_{t-1}$.}
\label{fig:inter-trajectory-attention}
\end{figure}

\begin{figure}[htbp!]
\centering
\includegraphics[width=0.95\linewidth]{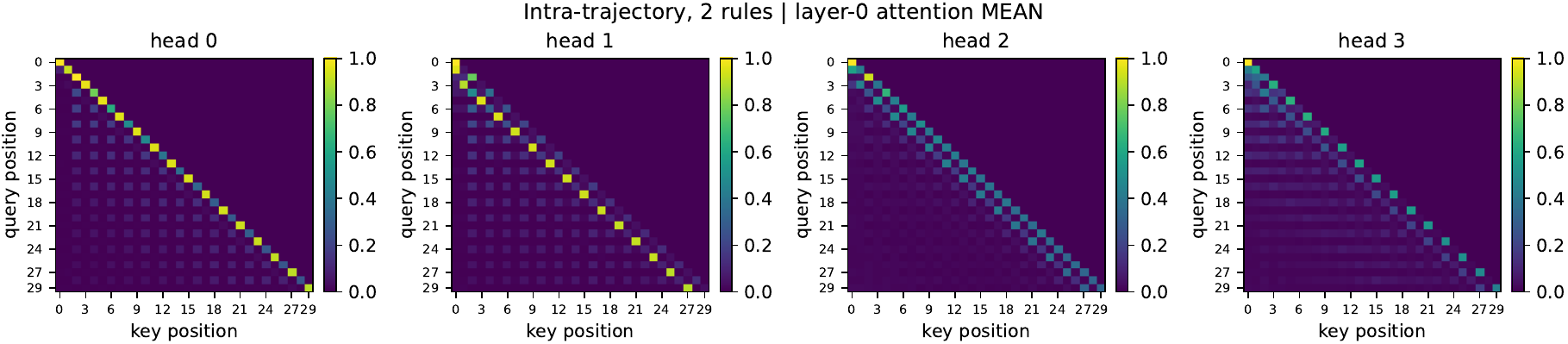}
\caption{Attention scores in the unmasked two-rule intra-trajectory setting. The model is trained without input masking.}
\label{fig:intra-trajectory-attention}
\end{figure}

\begin{figure}[htbp!]
\centering
\includegraphics[width=0.95\linewidth]{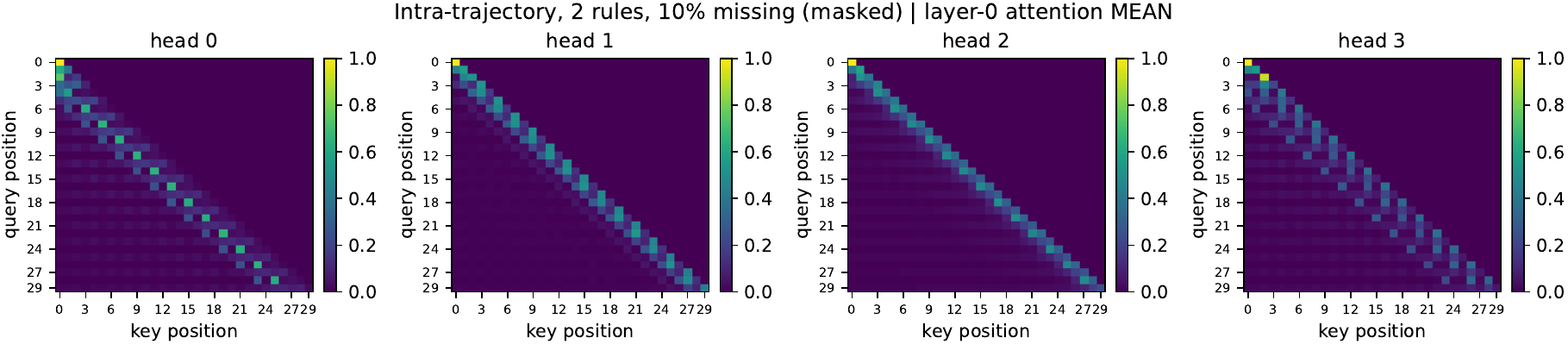}
\caption{Attention scores in the two-rule intra-trajectory setting with $10\%$ random input masking during training.}
\label{fig:intra-trajectory-attention-mask}
\end{figure}

We visualize the attention weights after softmax for the inter- and intra-trajectory settings. Figures~\ref{fig:inter-trajectory-attention}--\ref{fig:intra-trajectory-attention-mask} show the attention patterns for the unmasked inter-trajectory, unmasked intra-trajectory, and masked intra-trajectory settings, respectively. In all three settings, attention is not limited to the recent values needed by the local recurrence; earlier tokens can also receive nonzero attention. However, nonzero attention does not imply that a token is needed for prediction. We therefore use these figures only to show the attention patterns and rely on the intervention experiments in Section~\ref{sec:local_recurrence} to test the local recurrence hypothesis.

%% file: appendix_related_work.tex
\section{Related Work}
\label{sec:related}

\paragraph{Controlled algorithmic learning and mechanism study.}
Controlled algorithmic tasks allow predictive success to be compared with
the computations learned by a model. Modular arithmetic has exposed delayed
generalization and structured arithmetic algorithms
\citep{power2022grokking,nanda2023progress,zhong2023clock,li2025fourier},
while synthetic sequence tasks have enabled analyses of finite-state
computation, internal world representations, and prediction under partial
observability
\citep{liu2023automata,zhang2025finite,li2023world,shai2024belief,lu2024pomdp}.
Particularly close to our setting, \citet{he2024learningtogrok} study multiple
modular-linear functions $z=ax+by\bmod p$, including generalization to unseen
functions and the algorithms learned from input--output demonstrations.
Our tasks couple successive predictions through a recurrence and vary
whether the active rule is latent or explicitly supplied, as well as which
past values remain observable. 

\paragraph{Mechanistic interpretability of language models.}
Circuit analyses identify internal components that contribute to specific
language-model behaviors \citep{wang2023wild}, while studies of superposition
explain how multiple features can share representational dimensions
\citep{elhage2022superposition}. Sparse autoencoders have been used to extract
interpretable features from language-model activations
\citep{bricken2023monosemanticity}, with subsequent work addressing larger
models, evaluation, and open resources
\citep{templeton2024scaling,gao2025sae,lieberum2024gemmascope}.
Transcoders and cross-layer transcoders extend this analysis toward
interactions between features and computational graphs
\citep{dunefsky2024transcoders,ameisen2025tracing}, enabling case studies of
computations in deployed language models \citep{lindsey2025biology}.
Our controlled recurrence setting makes candidate rules explicit. We combine
linear probes with designed input perturbations to characterize rule
information in activations and test candidate prediction strategies.

\paragraph{In-context learning of tasks and dynamics.}
In-context learning has been studied through latent-concept inference
\citep{xie2022implicit}, learning simple function classes from examples
\citep{garg2022transformers}, and induction-head mechanisms
\citep{olsson2022context}. Analyses of linear regression connect attention
to optimization algorithms
\citep{vonoswald2023gradient,akyurek2023algorithm}, while task- and
function-vector studies investigate compact representations that condition
subsequent computation
\citep{hendel2023taskvectors,todd2024functionvectors,han2025taskvectors}.
For temporal data, prior work analyzes autoregressive prediction
\citep{sander2024autoregressive}, approximation and depth
\citep{cole2025lds}, and system identification
\citep{vladarean2026lds} for linear dynamics.
Their theoretical results assume restricted linear-attention models, while
our experiments use softmax Transformers to study discrete modular recurrences.


\paragraph{Multi-task learning and training curricula.}
Joint learning can improve generalization through shared representations
\citep{caruana1997multitask}, but can also introduce optimization interference
\citep{yu2020gradient}. Curriculum learning changes the order and difficulty
of training examples \citep{bengio2009curriculum}, while task diversity can
change the form of in-context generalization \citep{raventos2023diversity}.
We study these questions within a common recurrence family by varying
per-rule data fractions and progressively expanding the rule pool while
retaining earlier tasks. The resulting interference, transfer, and
curriculum effects characterize which rules are learned under the evaluated
training protocols, complementing our analysis of representational capacity.

%% file: experiment_details.tex
\section{Data Construction} 
\label{sec:exp_detail}
\subsection{Trajectory-Aware Data Split}
\label{sec:data_split}

A naive random split of sequence windows can introduce substantial train--test
leakage. Different windows sampled from the same recurrence orbit may share
states or local transitions, so randomly assigning these windows to training
and test sets could allow test-time transitions to have already appeared in
training.

To minimize this leakage, we first enumerate the recurrence orbits and then
generate examples by sliding a fixed-length window along each orbit, advancing
the window by one state at a time. We assign complete orbits to either the
training or test set whenever possible while targeting a $70\%/30\%$ split of
the state space.

Because orbit sizes do not generally align exactly with the desired ratio, at
most one orbit is split between the two sets. When such a split is necessary,
the orbit is divided contiguously: an initial portion is assigned to training
and the remaining portion to testing, rather than interleaving states from the
same orbit across the two sets. Thus, even in the extreme case where most
nonzero states lie on a single orbit, the test portion forms a contiguous
held-out segment whose internal state transitions are not exposed during
training, apart from the unavoidable boundary between the two portions.

This construction preserves the trajectory structure as much as possible under
the fixed train--test ratio and limits train--test leakage to at most a single
orbit boundary, instead of allowing shared local transitions throughout the
state space.

\subsection{Masking}
\label{sec:masking_train}

Masking replaces selected sequence values with a special token $\bot$ while preserving their positions; rule tags in the intra-trajectory scenario are never masked.

Masks are generated sequentially from left to right. At each eligible position,
a masked span is initiated with probability $p_{\mathrm{mask}}$. If a span is
initiated, its length is sampled as
$
l\sim\mathrm{Unif}\{1,\ldots,\texttt{maxMaskLen}\}.
$
The corresponding $l$ consecutive values are masked. The scan then resumes
after the masked span, with at least one unmasked position enforced before
another masked span may begin. This prevents adjacent spans from merging and
ensures that every contiguous masked region has length at most
\texttt{maxMaskLen}.
A contiguous span of masked values is called a \emph{hole}. We characterize a
sequence by its maximum hole length $m$ and refer to sequences with maximum hole
length $m$ as belonging to the \emph{misslen-$m$} group.
We use the maximum hole length as the primary control variable for masking
difficulty. The motivation is that longer consecutive missing spans may require
information to be propagated through more intermediate recursive steps before
a later value can be determined. We therefore study how Transformer capability,
and in particular the model depth required for successful prediction, changes
as the maximum hole length increases.

%% file: shortcut_details.tex
\section{Shortcut Definition and Existence}
\label{app:shortcuts}

\subsection{Definition of Shortcuts}
\label{app:shortcut_definition}

Let $\mathbb{F}_p$ denote the finite field with $p$
elements. Consider an order-$k$ linear recurrence
\begin{equation}
    q_t=\sum_{i=1}^{k}a_iq_{t-i}\pmod p,
    \label{eq:original_recurrence}
\end{equation}
where
$\mathbf a=(a_1,\ldots,a_k)\in\mathbb F_p^k$
and $a_k\neq0$.

A \emph{shortcut} is an alternative recurrence
that uses fewer than $k$ historical values.
Formally, it is specified by a coefficient vector
$\mathbf a'=(a'_1,\ldots,a'_{k'})
\in\mathbb F_p^{k'}$,
with $a'_{k'}\neq0$, such that every sequence
generated by the original recurrence also satisfies
\begin{equation}
    q_t=\sum_{i=1}^{k'}a'_iq_{t-i}\pmod p,
    \qquad t>k'.
    \label{eq:shortcut_recurrence}
\end{equation}

Define the set of active lags as
\begin{equation}
    \operatorname{supp}(\mathbf a')
    =\{i\in\{1,\ldots,k'\}:a'_i\neq0\}.
\end{equation}
The shortcut must satisfy
\begin{equation}
    |\operatorname{supp}(\mathbf a')|<k.
    \label{eq:shortcut_condition}
\end{equation}
A shortcut with exactly $n$ active lags is called
an \emph{$n$-term shortcut}.

Here, $k'$ specifies the furthest historical
position required by the shortcut.
For shortcuts with $k'>k$, the \emph{delay}
is defined as
\begin{equation}
    \mathrm{delay}=k'-k.
    \label{eq:shortcut_delay}
\end{equation}
This corresponds to the number of initial
predictions for which the shortcut cannot
yet be applied.

For example, the order-$3$ recurrence
\begin{equation}
    q_t=2q_{t-1}+6q_{t-2}+9q_{t-3}
    \pmod{127}
\end{equation}
admits a two-term shortcut
\begin{equation}
    q_t=16q_{t-2}+81q_{t-6}
    \pmod{127}.
    \label{eq:experimental_shortcut}
\end{equation}
This shortcut has active lags $\{2,6\}$
and a delay of $6-3=3$.

\subsection{Existence of Shortcuts}
\label{app:shortcut_existence}

\textbf{Periodicity provides a trivial shortcut.}
Every invertible linear recurrence over a finite
field admits a shortcut due to its periodicity.
Consider an order-$k$ recurrence
\begin{equation}
    q_t=\sum_{i=1}^{k}a_iq_{t-i}\pmod p,
\end{equation}
where $k\geq2$ and $a_k\neq0$.
Let $s_t=(q_t,\ldots,q_{t+k-1})^\top$ denote
its state vector, with state transition
$s_{t+1}=As_t$.
Since $\det A=(-1)^{k-1}a_k\neq0$,
the transition matrix $A$ is invertible.

As there are finitely many invertible matrices
over $\mathbb F_p$, there exists a positive
integer $T$ such that $A^T=I$.
Consequently, every sequence generated by the
recurrence satisfies $q_t=q_{t-T}$ whenever
both terms are defined.
Choosing a positive multiple $M>k$ of $T$
gives
\begin{equation}
    q_t=q_{t-M}, \qquad t>M.
    \label{eq:trivial_shortcut}
\end{equation}

\textbf{The existence of general shortcuts is guaranteed by algebraic properties.}
For an order-$k$ recurrence, the existence of shortcuts
can be characterized algebraically using its characteristic
polynomial. By representing polynomial remainders as
vectors over a finite field, the shortcut condition
can be expressed as a linear dependence condition.
The pigeonhole principle then guarantees that such
a dependence occurs, establishing the existence
of shortcuts for arbitrary recurrence orders.

%% file: shortcut_verify.tex
\section{Identifying Recurrence Formulas Learned by Transformers}
\label{app:shortcut_identification}

\subsection{Experimental Setup}
\label{app:how_rule_fit}

To identify the recurrence formulas learned by
Transformers, we combine attention analysis with
controlled input interventions.
This approach allows us to investigate both
alternative formulas used under masked observations (Appendix~\ref{sec:rule_fit_masking})
and shortcuts learned in higher-order tasks (Appendix~\ref{sec:rule_fit_trib}).
Our method consists of two stages: identifying
candidate historical positions from attention
patterns and recovering the algebraic relationships
between these positions and the model's predictions.

The central idea is to construct intervention inputs
that do not necessarily satisfy the original recurrence.
In valid recurrence sequences, historical values are
algebraically dependent, so multiple equivalent formulas
may produce identical predictions.
Consequently, prediction accuracy on valid sequences
alone cannot reveal which historical values the model
actually uses.
By independently varying the input values, we break
these dependencies and examine the model's
input-output relationship.

\paragraph{Step 1: Identifying candidate positions
through attention analysis.}

We first visualize the attention matrices across
different layers and heads.
For each prediction position, we collect the historical
positions receiving substantial attention and express
them as relative distances from the current position.

We aggregate these attention patterns across heads
and layers to identify candidate historical positions.
Consecutive prediction positions with similar attention
patterns are grouped together for subsequent analysis.

Importantly, attention patterns only indicate where
the model retrieves information; they do not establish
how the retrieved values contribute to its prediction.
We therefore treat the attended positions as candidates
rather than direct evidence of a particular recurrence
formula.

\paragraph{Step 2: Recovering algebraic relationships
through off-manifold probing.}

To identify the computation performed by the model,
we construct artificial input sequences in which
the observed values are sampled independently and
uniformly from $\mathbb F_p$.
Unlike the original training sequences, these inputs
are not constrained to satisfy any recurrence relation.
For masked experiments, we preserve the mask pattern,
represent masked positions by $\bot$, and independently
sample only the unmasked values.

Independent random inputs remove the algebraic
dependencies among historical values, allowing
different candidate formulas to be distinguished
through their predictions.

For a model predicting $q_{t+1}$, let $D_t$ denote
the set of candidate attention distances,
augmented with the three most recent positions
used by the original order-$3$ recurrence.
For masked inputs, we retain only the candidate
positions that are observable under the given
mask pattern.

We examine whether the model's predictions
on the artificial inputs satisfy a linear
relationship of the form
\begin{equation}
    \hat{q}_{t+1}
    \equiv
    \sum_{d\in D_t}\beta_d q_{t-d}
    \pmod p,
    \label{eq:shortcut_probe_regression}
\end{equation}
where $\beta_d\in\mathbb F_p$ are the
coefficients to be recovered.

We collect the model's predictions on independently
generated input sequences and fit the coefficients
over $\mathbb F_p$.
We first attempt to solve the resulting system
exactly using modular Gaussian elimination.
If no exact solution exists, we search for
a best-fitting linear formula using RANSAC.

To quantify the quality of a recovered formula,
we calculate its prediction agreement:
\begin{equation}
    \operatorname{Agreement}
    =
    \frac{1}{|\mathcal D|}
    \sum_{\mathbf q\in\mathcal D}
    \mathbb{I}
    \left[
        \hat{q}_{t+1}(\mathbf q)
        =
        \left(
        \sum_{d\in D_t}\beta_d q_{t-d}
        \right)\bmod p
    \right],
    \label{eq:shortcut_probe_agreement}
\end{equation}
where $\mathcal D$ is the collection of
artificial probe inputs.

The recovered nonzero coefficients identify
the historical positions contributing to the
fitted formula.
We compare the recovered formulas with the original
recurrence and examine how the models use different
historical positions in the presence or absence
of masking.

Because the probe inputs violate the original
recurrence constraints, high agreement cannot
be attributed solely to algebraic correlations
in the training distribution.
This provides stronger evidence of the model's
computational behavior than attention visualization
or prediction accuracy on valid sequences alone.

\subsection{Identifying Alternative Formulas Under Masked Single-rule Recurrence}
\label{sec:rule_fit_masking}

We apply our rule-identification method to the
single-rule setting with masked observations to investigate
how models handle missing historical values.

\begin{figure}[t]
    \centering
    \includegraphics[width=0.8\linewidth]{
        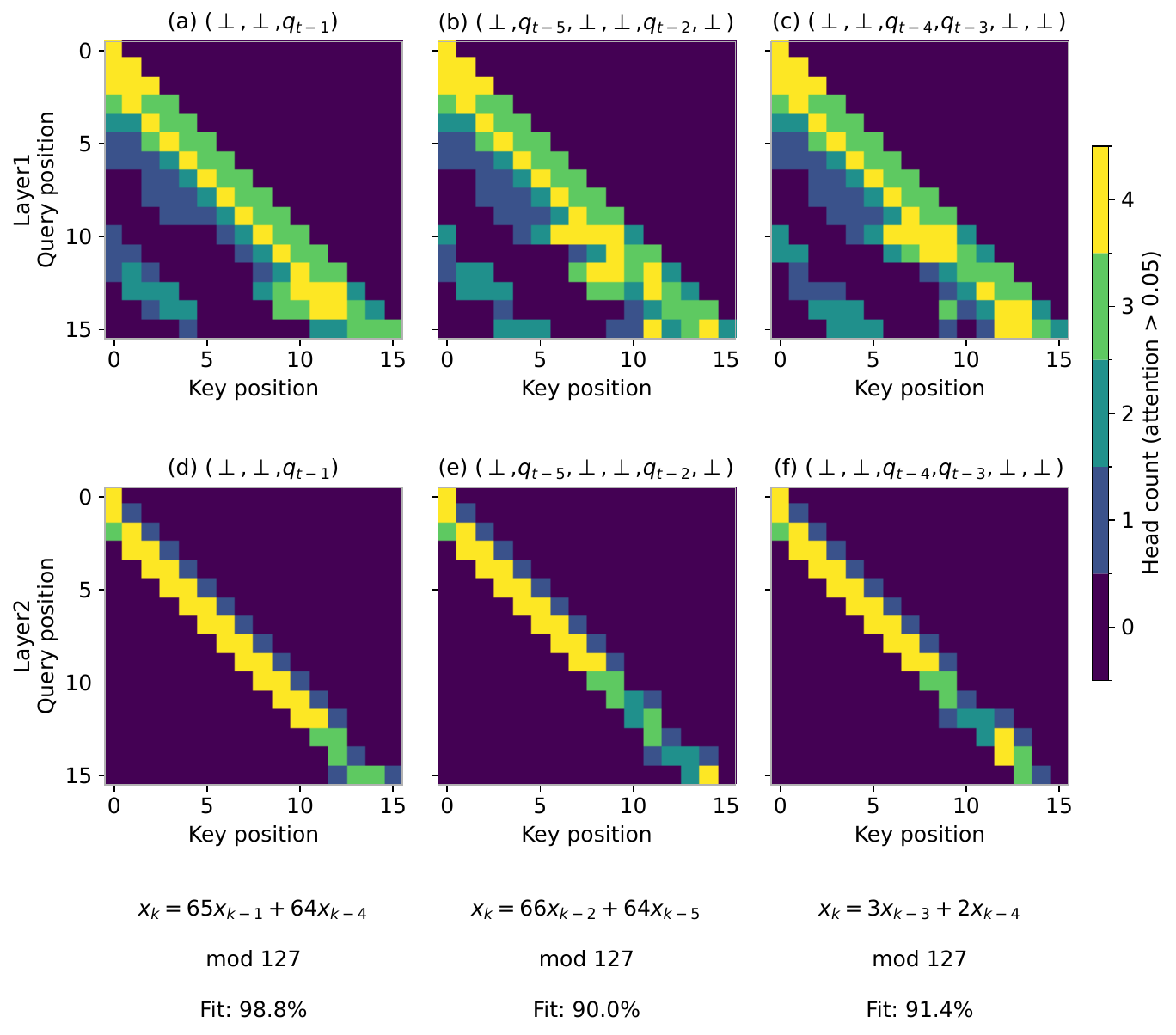
    }
    \caption{
    Attention patterns and recovered recurrence formulas
    under three representative masking patterns in the
    single-rule setting.
    Each column corresponds to a different masking pattern,
    where $\bot$ denotes a masked token and $q$ an observed value.
    The first two rows show attention maps for the two
    Transformer layers. Each cell counts the number of
    heads (out of four) whose attention weight exceeds $0.05$.
    The bottom row reports the recurrence formula recovered
    through off-manifold probing and its prediction agreement.
    All three patterns yield alternative formulas distinct
    from the original $(1,1)$ recurrence, illustrating how
    the model exploits different historical positions
    to predict under missing observations.
    }
    \label{fig:masked_rule_fit}
\end{figure}

Figure~\ref{fig:masked_rule_fit} presents three representative
masking patterns: $(\bot,\bot,q_{t-1})$,
$(\bot,q_{t-5},\bot,\bot,q_{t-2},\bot)$, and
$(\bot,\bot,q_{t-4},q_{t-3},\bot,\bot)$,
where $\bot$ denotes a masked token and $q$ an observed
value. For each pattern, attention analysis
identifies candidate historical positions, while
off-manifold probing recovers the corresponding
prediction formula.

Although the original recurrence uses the two
immediately preceding values with coefficients $(1,1)$,
all three recovered formulas involve alternative
historical positions and coefficients.
In particular, the third pattern masks both immediately
preceding values, yet the model successfully predicts
using values from more distant positions.
The recovered formulas achieve prediction agreement
ranging from $90.0\%$ to $98.8\%$ on the probing inputs.

These results provide evidence that Transformers can
learn multiple algebraically equivalent recurrence
formulas and select alternative formulas based on
the available historical values under different
masking patterns.
This helps explain how increasing model width
can improve generalization under masking, as observed
in Fig.~\ref{fig:single_mask_width}.

\subsection{Identifying Shortcuts in Order-$3$ Tasks}
\label{sec:rule_fit_trib}

We apply the procedure above to four representative
models, selected to illustrate how shortcut delay
and loss reweighting affect the recurrence formulas
used at different prediction positions.

\begin{figure}[htbp]
    \centering
    \includegraphics[width=\linewidth]{
        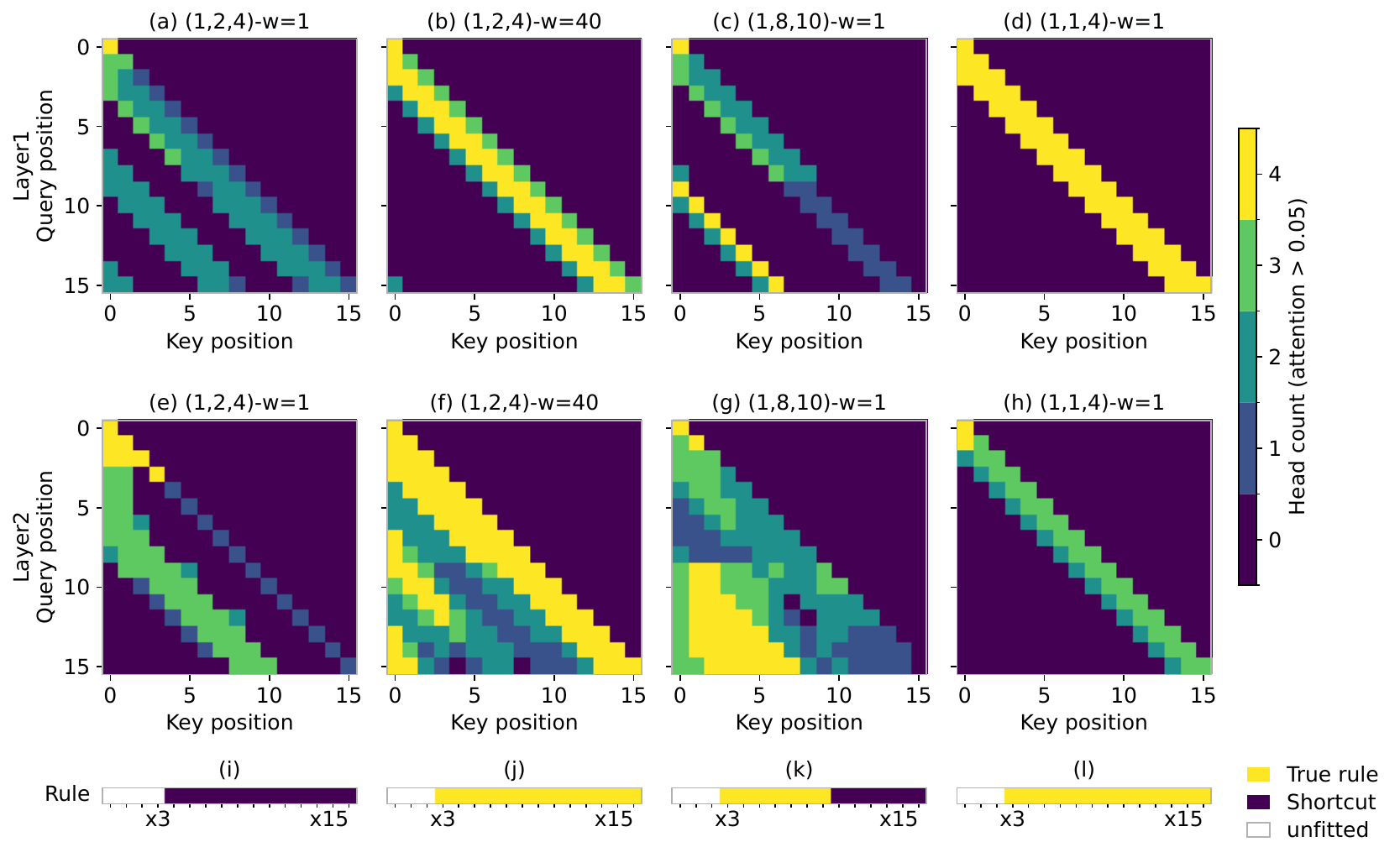
    }
    \caption{
    Attention patterns and recovered recurrence rules for four
    representative order-$3$ tasks.
    Each column corresponds to one experiment: shortcut delays
    of 1, 1, 9, and 13, respectively, with the second experiment
    using a first-prediction loss weight of $w=40$
    ($w=1$ otherwise).
    The first and second rows show attention maps for the two
    Transformer layers. Each cell counts the number of
    attention heads (out of four) whose attention weight exceeds
    $0.05$ for the corresponding query and key positions.
    The bottom row classifies the formula recovered at each
    prediction position through off-manifold probing as the
    original recurrence (yellow), a shortcut (purple),
    or without a reliable fit (white).
    }
    \label{fig:rule_fit_trib}
\end{figure}

\emph{Small delay} (Figure~\ref{fig:rule_fit_trib}(a)).
In the first example, the recovered formulas
are consistent with the original order-$3$
recurrence at the earliest prediction positions.
Once sufficient historical context becomes
available, the recovered formula changes
to a two-term shortcut.
This example illustrates how a model can
use different algebraically equivalent formulas
at different positions within the same sequence.

\emph{Intermediate delay} (Figure~\ref{fig:rule_fit_trib}(c)).
We observe a similar transition in the
second example, but the region in which
the model uses the original recurrence
extends over more prediction positions.
The recovered formula changes to a shortcut
only at later positions, consistent with
the longer context required by its shortcut delay.

\emph{Large delay} (Figure~\ref{fig:rule_fit_trib}(d)).
In contrast, the third example shows no
transition to a shortcut within the evaluated
prediction positions.
The recovered formulas remain consistent
with the original order-$3$ recurrence
throughout the sequence.
This behavior illustrates how a sufficiently
long shortcut delay can favor learning
the original recurrence within the
available context.

\emph{Loss reweighting} (Figure~\ref{fig:rule_fit_trib}(b)).
Finally, we revisit the short-delay example
after increasing the loss weight factor of the first
prediction position to 40.
Unlike the corresponding model trained
with the standard loss, the recovered
formulas remain consistent with the
original recurrence across all evaluated
prediction positions.
Thus, although a short shortcut is
algebraically available, the training
objective can discourage the model
from relying on it.

Together, these examples show that shortcut
availability alone does not determine
which recurrence formula the model learns.
The formula used at each prediction position
also depends on the training objective.

%% file: failed_tasks.tex
\section{Additional Experiments on Challenging Tasks}
\label{app:fail}

This section presents additional experiments on
challenging task configurations, including multi-rule
learning, masked observations, and higher-order
recurrences. We examine the effects of increasing
model capacity and investigate how the availability
of shortcuts affects learning.

\subsection{Four-Rule Inter-trajectory Learning}
\label{app:fail_N4}

\begin{figure}[h]
    \centering
    \includegraphics[width=\linewidth]
    {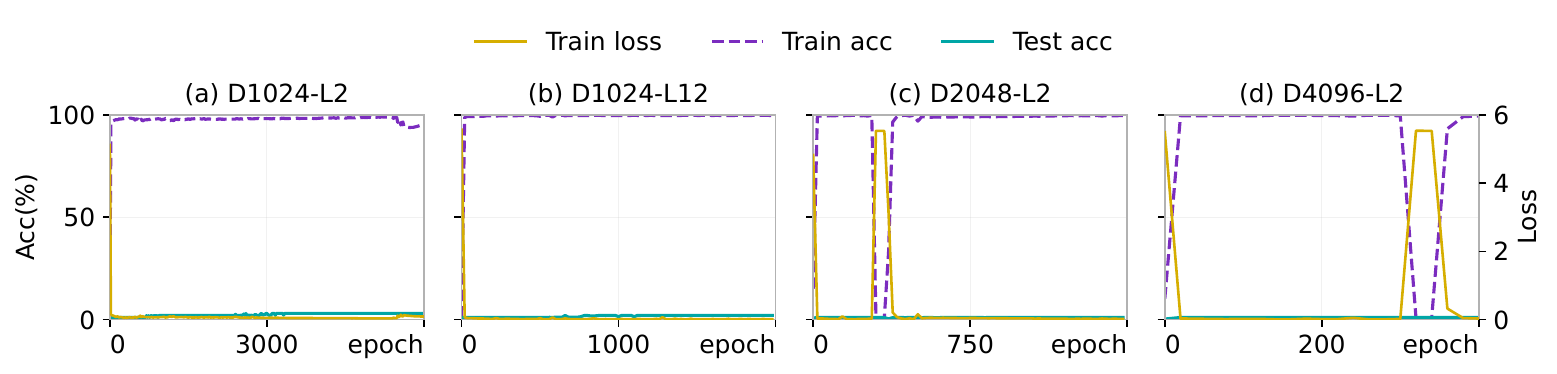}
    \caption{
    Joint training on four-rule inter-trajectory tasks
    with different model depths and widths.
    Panels (a)--(d) correspond to
    $(d_{\mathrm{model}},L) =
    (1024,2), (1024,12), (2048,2), (4096,2)$,
    respectively.
    Each panel reports training loss, training
    accuracy, and test accuracy.
    All four configurations achieve high training
    accuracy but fail to generalize within their
    respective training budgets.
    }
    \label{fig:fail_N4}
\end{figure}

We investigate whether increasing model depth or width
enables successful joint learning of four latent rules.
As shown in Figure~\ref{fig:fail_N4}, neither increasing
depth to 12 layers nor increasing width to 4096
leads to successful generalization within the tested
training budgets.
Although the models achieve high training accuracy,
their test accuracy remains near chance.
These results contrast with curriculum learning,
which enables models to acquire additional latent
rules by progressively introducing them, as
discussed in Section~\ref{sec:curriculum}.

\subsection{Masked Inter-trajectory Learning}
\label{app:fail_mixed_mask}

\begin{figure}[h]
    \centering
    \includegraphics[width=\linewidth]
    {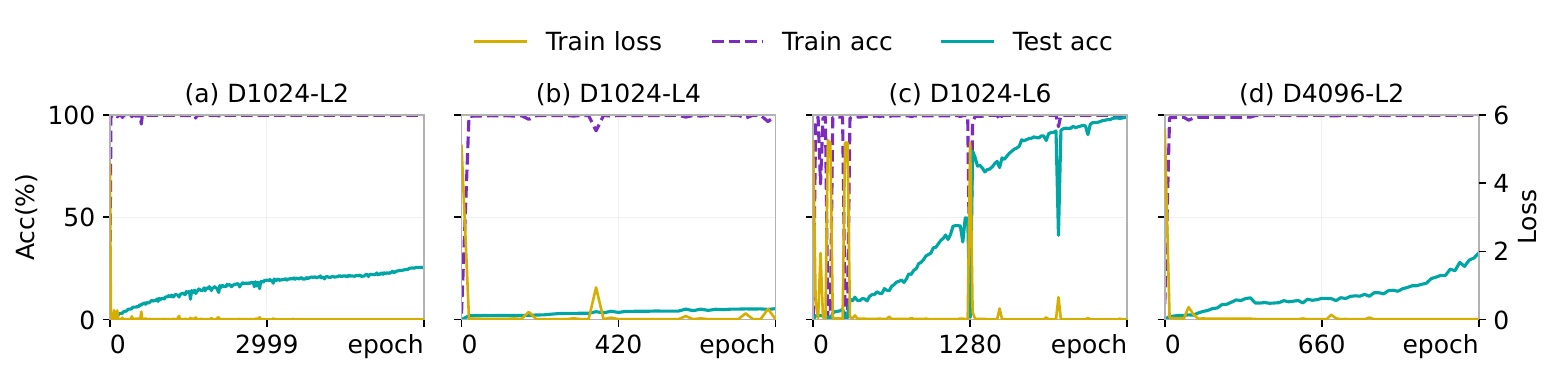}
    \caption{
    Learning curves for the masked two-rule
    inter-trajectory setting.
    Panels (a)--(d) correspond to
    $(d_{\mathrm{model}},L) =
    (1024,2), (1024,4), (1024,6), (4096,2)$,
    respectively.
    Only the six-layer model in panel (c)
    achieves high test accuracy, while the
    other configurations fail to generalize
    within their respective training budgets.
    Each panel reports training loss, training
    accuracy, and test accuracy.
    }
    \label{fig:fail_mixed_mask}
\end{figure}

We further investigate the combination of latent-rule
learning and incomplete observations.
Figure~\ref{fig:fail_mixed_mask} shows that this
setting requires substantially greater model capacity.
Among the four tested configurations, only the
six-layer model successfully generalizes.
Notably, increasing depth from two to six layers
enables successful learning, whereas increasing
width from 1024 to 4096 does not achieve the
same result within the tested training budgets.
This suggests that depth and width play different
roles in learning masked latent-rule tasks.

We hypothesize that masking further complicates
latent-rule identification by removing observations
that would otherwise help distinguish different
recurrence rules.

\subsection{Challenging Order-4 Recurrences}
\label{app:fail_order_4}

\begin{figure}[h]
    \centering
    \includegraphics[width=\linewidth]{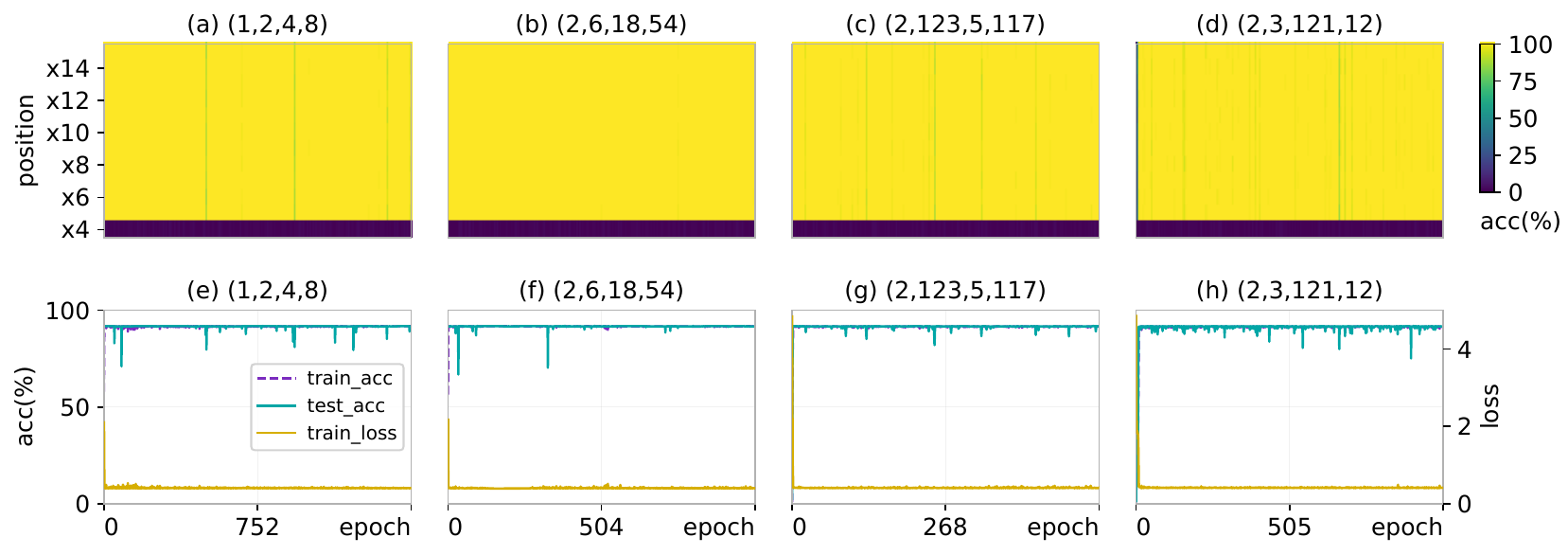}
\caption{
Order-$4$ recurrences
with two-term shortcuts ($p=127$,
$d_{\mathrm{model}}=1024$, $L=2$).
From left to right, the learned shortcuts are
$q_t=2q_{t-1}-q_{t-5}$,
$q_t=5q_{t-1}+92q_{t-5}$,
$q_t=124q_{t-3}+107q_{t-5}$, and
$q_t=7q_{t-2}+24q_{t-5}$,
all modulo $127$.
The top row shows per-position test accuracy
over training epochs, while the bottom row
shows the corresponding training accuracy,
test accuracy, and training loss. }
    \label{fig:order4_success}
\end{figure}

\begin{figure}[h]
    \centering
    \includegraphics[width=\linewidth]
    {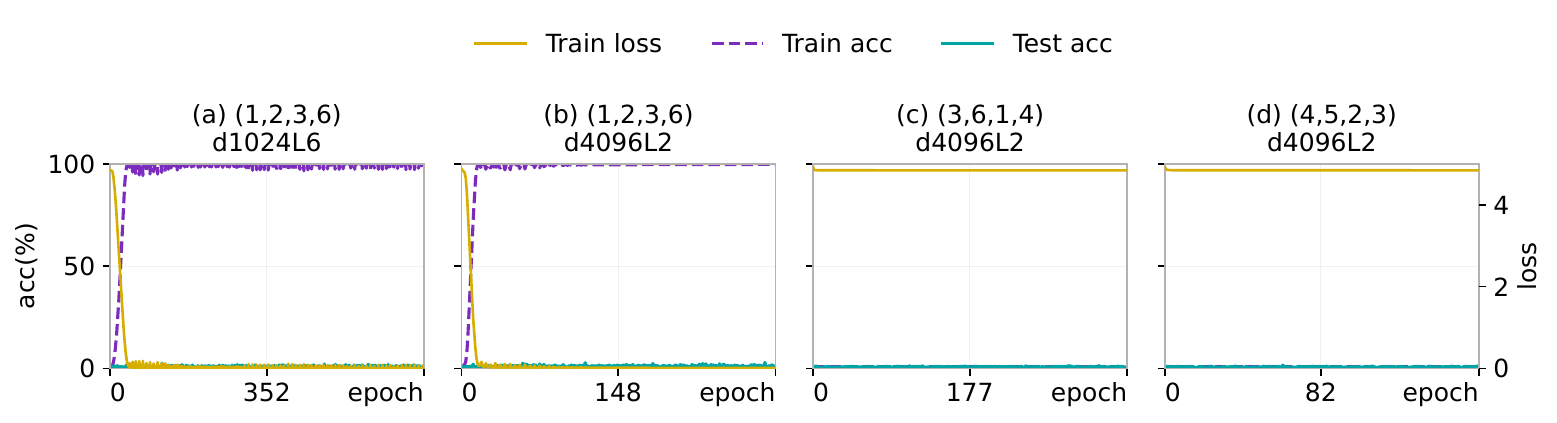}
    \caption{
    Learning curves for challenging order-$4$
    recurrences over $\mathbb{F}_{127}$.
    Panels (a) and (b) examine the rule
    $(1,2,3,6)$ with configurations
    $(d_{\mathrm{model}},L)=(1024,6)$
    and $(4096,2)$, respectively.
    Panels (c) and (d) examine rules
    $(3,6,1,4)$ and $(4,5,2,3)$,
    respectively, using
    $(d_{\mathrm{model}},L)=(4096,2)$.
    The first two configurations achieve high
    training accuracy but fail to generalize,
    while the remaining two fail to fit the
    training data under their respective
    training configurations.
    Curves show training loss, training
    accuracy, and test accuracy.
    }
    \label{fig:fail_order4}
\end{figure}

Order-$4$ recurrences may admit both two-term and
three-term shortcuts.
Our experiments show that learning order-$4$
recurrences remains challenging.
We observe successful generalization only in cases
where models learn a two-term shortcut
(Figure~\ref{fig:order4_success}).
However, tasks beyond two-term shortcut are challenging and Transformers 
fail to generalize
within the training budget (Figure~\ref{fig:fail_order4}).

\subsection{Order-$3$ Intra-trajectory Learning}
\label{app:fail_3_intra}

We investigate order-$3$ recurrences in the
intra-trajectory setting, where the active rule
may change at every prediction step.
Unlike fixed-rule recurrences, these sequences
do not generally satisfy the shortcuts derived
for a single recurrence rule, since successive
values may be generated by different rules.

\begin{figure}[h]
    \centering
    \includegraphics[width=\linewidth]
    {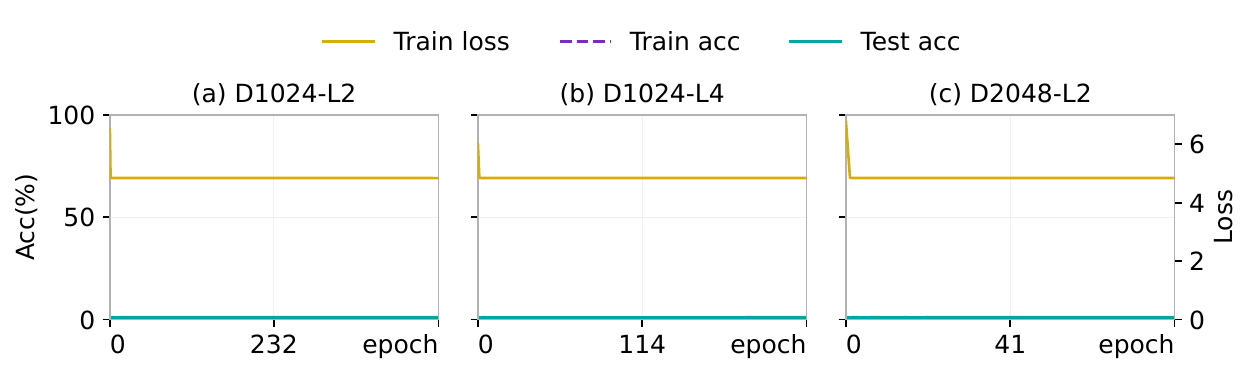}
    \caption{
    Learning curves for the two-rule,
    order-$3$ intra-trajectory setting.
    Panels (a)--(c) correspond to
    $(d_{\mathrm{model}},L) =
    (1024,2), (1024,4), (2048,2)$,
    respectively.
    All three configurations remain near
    chance test accuracy within their
    respective training budgets.
    Curves show training loss, training
    accuracy, and test accuracy.
    }
    \label{fig:fail_3_intra}
\end{figure}

Consequently, models cannot directly exploit
the fixed-rule shortcuts discussed in
Section~3.3.
Figure~\ref{fig:fail_3_intra} shows that all
three tested configurations fail to generalize.
Increasing either model depth or width does
not resolve this difficulty within the
evaluated training budgets.

\subsection{Order-$3$ Inter-trajectory Learning}
\label{app:fail_3_inter}

\begin{figure}[h]
    \centering
    \includegraphics[width=\linewidth]
    {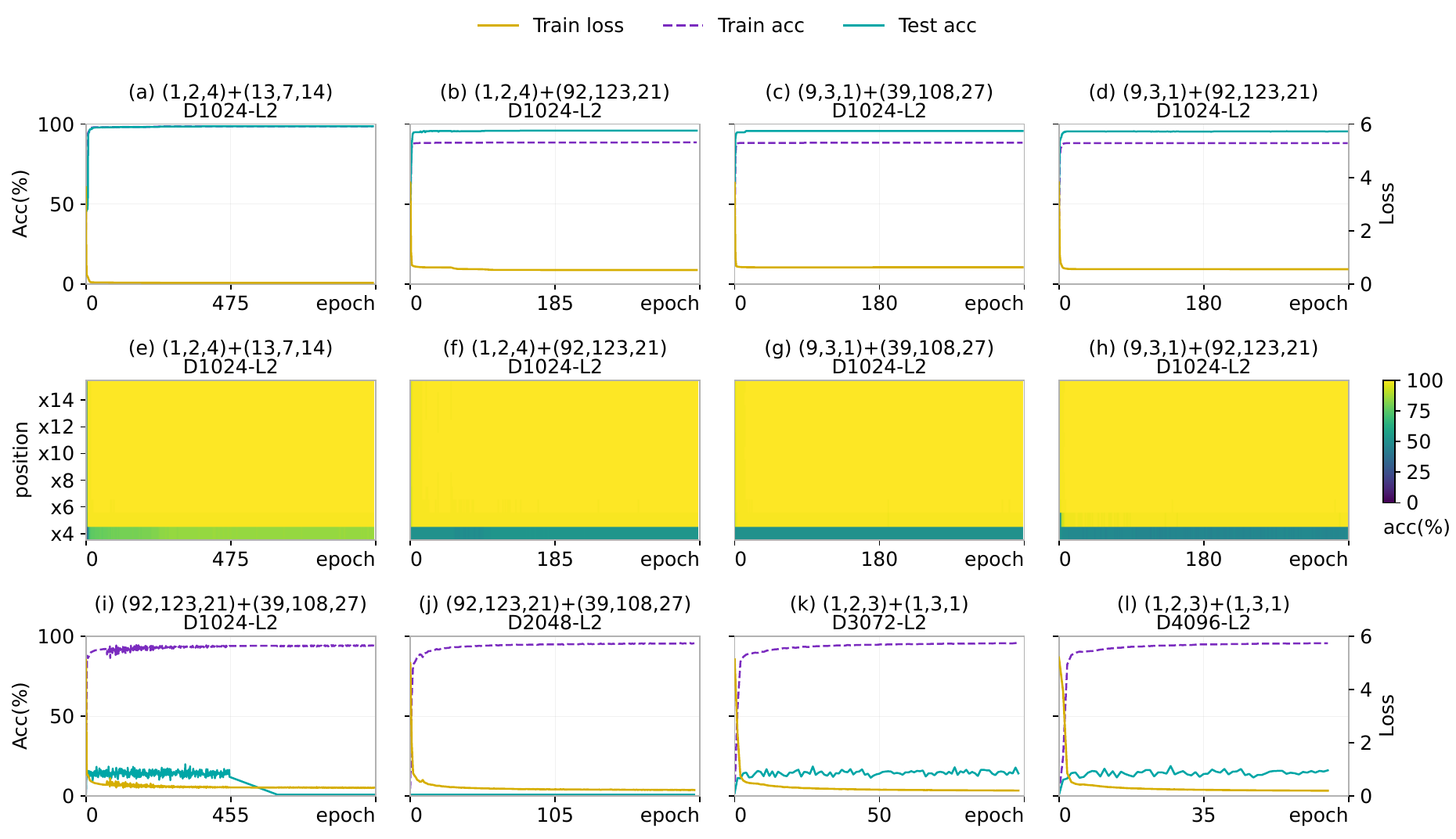}
    \caption{
    Order-$3$ inter-trajectory learning with
    different pairs of recurrence rules
    over $\mathbb{F}_{127}$.
    The top row shows learning curves for
    rule pairs
    $(1,2,4)+(13,7,14)$,
    $(1,2,4)+(92,123,21)$,
    $(9,3,1)+(39,108,27)$, and
    $(9,3,1)+(92,123,21)$, respectively,
    using $d_{\mathrm{model}}=1024$ and $L=2$.
    The middle row shows the corresponding
    per-position test accuracy.
    Although these configurations achieve
    high overall test accuracy by exploiting
    short-delay shortcuts, they fail at the
    first prediction position.
    The bottom row presents additional
    experiments with longer-delay shortcuts:
    $(92,123,21)+(39,108,27)$ with
    $(d_{\mathrm{model}},L)=(1024,2)$
    and $(2048,2)$, and
    $(1,2,3)+(1,3,1)$ with
    $(d_{\mathrm{model}},L)=(3072,2)$
    and $(4096,2)$.
    None of these configurations achieves
    successful generalization.
    Learning curves report training loss,
    training accuracy, and test accuracy.
    }
    \label{fig:fail_3_inter}
\end{figure}

Finally, we investigate the interaction between
latent-rule identification and higher-order
recurrence learning.
Figure~\ref{fig:fail_3_inter} compares several
pairs of order-$3$ rules with different
shortcut properties.

When at least one rule admits a shortcut with
delay $1$, models can achieve approximately
95\% overall test accuracy.
The per-position accuracy reveals that these
models generalize at later positions but fail
at the first prediction position, where the
shortcut is not yet applicable.
In contrast, when both rules have longer
shortcut delays, none of the tested configurations
achieves successful generalization.

These observations suggest that short-delay
shortcuts can facilitate higher-order multi-rule
learning, even when models fail to acquire
the original recurrence.

%% file: theory.tex
\section{Expressivity of two-layer causal Transformers}
\label{app:theory}

We give an explicit construction for the unmasked inter-trajectory setting
of Section~\ref{sec:model_training}. A two-layer causal Transformer can predict the
next value for every initial state and every rule in a fixed pool, once it
has observed $2k$ values. The construction first checks each candidate rule
at individual positions and then combines the most recent checks. We
specify the architecture, count its resources, and prove correctness for
ordinary softmax attention with finite parameters.


\subsection{Task and precise causal Transformer model}
\label{app:theory:model}

\paragraph{Task and quantifiers.}
We retain the notations from
Section~\ref{sec:setup}, and write the fixed pool of $K$ rules as
\[
 \mathcal{A}=\{a^{(1)},\ldots,a^{(K)}\}\subseteq \mathbb{F}_p^k,
 \qquad a^{(z)}=(a^{(z)}_1,\ldots,a^{(z)}_k).
\]
Here $p\ge2$ is prime, $k\ge1$ is the order of the recurrence, and $1\le K\le p^k$ is the number of rules. We suppress the
trajectory index $j$. A legal trajectory is generated by one fixed,
unobserved $z\in\{1,\ldots,K\}$ and arbitrary initial values
$q_1,\ldots,q_k\in \mathbb{F}_p$, according to
\begin{equation}
 q_t=\sum_{i=1}^k a^{(z)}_i q_{t-i}\pmod p,
 \qquad t\ge k+1.
 \label{eq:theory:recurrence}
\end{equation}
Field elements and coefficients are represented by integers in
$\{0,\ldots,p-1\}$. 
All network computations take place over the real numbers. We also denote $N$ as the length of the $q$ sequence.



\paragraph{Architecture and dimensions.}
 We use $L$ to denote the layer of the Transformer, each
containing causal multi-head self-attention and a position-wise
feed-forward network, with residual connections around both sublayers.
For brevity, write $d=d_{\mathrm{model}}$ and $d_h=d/H$. We require
$d_h$ to be an even integer. The network uses ReLU activation,
omits normalization in the blocks and at the output. We use ReLU in the construction in order to simplify the construction. Our results also hold for GELU functions used in our experiments, see appendix \ref{app:theory:gelu} for details.

Each value token has a learned embedding in $\mathbb{R}^d$. We choose
the embedding table to realize
\begin{equation}
 h_t^{(0)}=(1,q_t,0,\ldots,0)^{\mathsf T}\in\mathbb{R}^d.
 \label{eq:theory:embedding}
\end{equation}
The first coordinate is a shared constant, independent of position. There
are no additive positional features. In block $\ell$, head $h$ has
matrices $W_{Q,h}^{(\ell)},W_{K,h}^{(\ell)},W_{V,h}^{(\ell)}
\in\mathbb{R}^{d_h\times d}$. Define
\begin{align}
 \mathbf{Q}_{t,h}^{(\ell)}&=W_{Q,h}^{(\ell)}h_t^{(\ell-1)},&
 \mathbf{K}_{t,h}^{(\ell)}&=W_{K,h}^{(\ell)}h_t^{(\ell-1)},&
 \mathbf{V}_{t,h}^{(\ell)}&=W_{V,h}^{(\ell)}h_t^{(\ell-1)}.
 \label{eq:theory:qkv}
\end{align}
The bold query and key vectors are distinct from the scalar trajectory
values $q_t$ and pool size $K$.

\paragraph{RoPE and the causal mask.}
Let
\begin{equation}
 \begin{split}
 \throt(u)&=\begin{pmatrix}\cos u&-\sin u\\\sin u&\cos u\end{pmatrix},\\
 R_t&=\operatorname{diag}\bigl(\throt(t\omega_0),\ldots,
                \throt(t\omega_{d_h/2-1})\bigr),
 \qquad \omega_b=10000^{-2b/d_h}.
 \end{split}
 \label{eq:theory:rope}
\end{equation}
In particular, $\omega_0=1$. 
For $1\le s\le t$, the attention scores, weights, and head outputs are
\begin{align}
 A_{t,s,h}^{(\ell)}
 &=\frac{(R_t\mathbf{Q}_{t,h}^{(\ell)})^{\mathsf T}
                (R_s\mathbf{K}_{s,h}^{(\ell)})}{\sqrt{d_h}},
 \label{eq:theory:scores}\\
 \alpha_{t,s,h}^{(\ell)}
 &=\frac{\exp A_{t,s,h}^{(\ell)}}
         {\sum_{u=1}^{t}\exp A_{t,u,h}^{(\ell)}},
 &o_{t,h}^{(\ell)}&=\sum_{s=1}^{t}
       \alpha_{t,s,h}^{(\ell)}\mathbf{V}_{s,h}^{(\ell)}.
 \label{eq:theory:attention}
\end{align}
Note that positions $s>t$ are excluded by the causal mask.

Define $\threlu(u)=\max\{u,0\}$ acting coordinatewise, an attention layer is
\begin{align}
 \widetilde h_t^{(\ell)}
 &=h_t^{(\ell-1)}+
   W_O^{(\ell)}\operatorname{concat}_{h=1}^{H}o_{t,h}^{(\ell)},
 \label{eq:theory:attention-residual}\\
 h_t^{(\ell)}
 &=\widetilde h_t^{(\ell)}+
   W_2^{(\ell)}\threlu\bigl(W_1^{(\ell)}\widetilde h_t^{(\ell)}
                              +b_1^{(\ell)}\bigr)+b_2^{(\ell)}.
 \label{eq:theory:ffn-residual}
\end{align}
Here $W_O^{(\ell)}\in\mathbb{R}^{d\times d}$,
$W_1^{(\ell)}\in\mathbb{R}^{w_\ell\times d}$,
$W_2^{(\ell)}\in\mathbb{R}^{d\times w_\ell}$,
$b_1^{(\ell)}\in\mathbb{R}^{w_\ell}$, and
$b_2^{(\ell)}\in\mathbb{R}^d$. Thus $w_\ell$ is the hidden width of a
single-hidden-layer feed-forward network (FFN), and
$d_{\mathrm{ff}}=\max_\ell w_\ell$. All maps are shared across positions;
weights may differ across blocks and heads. The final logits and prediction
are
\begin{equation}
 \lambda_v(q_{1:n})=(W_{\mathrm{out}}h_n^{(L)})_v,
 \quad v\in\{0,\ldots,p-1\},\qquad
 \widehat q_{n+1}=\argmax_{v\in\{0,\ldots,p-1\}}\lambda_v(q_{1:n}),
 \label{eq:theory:output}
\end{equation}
where $W_{\mathrm{out}}\in\mathbb{R}^{p\times d}$. 

\subsection{Expressiveness of the Transformer}
\label{app:theory:result}
Define the constants
\begin{equation}
 B=p-1,\qquad U=kB^2,\qquad J=\lfloor U/p\rfloor.
 \label{eq:theory:arithmetic-constants}
\end{equation}

\begin{theorytheorem}[Two-layer construction]
\label{thm:theory:two-layer}
For every fixed $(p,k,\mathcal{A})$ and every $N\ge2k$, there is a
Transformer of the class in Section~\ref{app:theory:model} with
\begin{equation}
 L=2,\qquad H=k,\qquad
 d_{\mathrm{model}}=2k\left\lceil\frac{kK+k+4}{2k}\right\rceil,
 \qquad d_{\mathrm{ff}}\le3K(J+1),
 \label{eq:theory:resources}
\end{equation}
such that, simultaneously for every legal prefix $q_{1:n}$ with
$2k\le n\le N$,
\begin{equation}
 \lambda_{q_{n+1}}(q_{1:n})-\lambda_v(q_{1:n})\ge\tfrac12
 \qquad\text{for every }v\ne q_{n+1}.
 \label{eq:theory:final-margin}
\end{equation}
All parameters are finite real numbers and independent of the active
rule, initial state, and actual prefix length. Consequently, the network
predicts the next token exactly, with residual width $O(kK+k)$,
per-head dimension $O(K+1)$, and FFN hidden width $O(Kkp)$.
\end{theorytheorem}

In the next sections, we provide the construction and prove the theorem.

\subsection{Continuation without unique rule identification}
\label{app:theory:continuation}

\begin{theorylemma}[Unique continuation from $2k$ values]
\label{lem:theory:continuation}
If two infinite sequences satisfy respective order-at-most-$k$ linear
recurrences over $C_p$ and agree on their first $2k$ values, then they
agree at every subsequent position.
\end{theorylemma}
\begin{theoryproof}
Let $T$ be the forward shift, $(Tq)_t=q_{t+1}$. The monic polynomial
\[
 P_a(X)=X^k-\sum_{i=1}^k a_iX^{k-i}
\]
annihilates a trajectory with rule $a$: $P_a(T)q=0$ at every index
$t\ge1$. For trajectories $q,q'$ with respective rules $a,a'$, the
polynomials in $T$ commute, so
\[
 P_a(T)P_{a'}(T)(q-q')=0.
\]
Their product is monic of degree $2k$. Hence every difference at position
$t+2k$ is a linear combination of the preceding $2k$ differences. Since
the first $2k$ differences vanish, induction gives $q-q'=0$ everywhere.
No nonzero constant coefficient or recurrence invertibility is needed.
\end{theoryproof}

\begin{theorycorollary}[Screening on the most recent window]
\label{cor:theory:screening}
For a legal prefix of length $n\ge2k$, let
\begin{equation}
 \mathcal{C}_n=\left\{z\in\{1,\ldots,K\}:
 q_t=\sum_{i=1}^k a^{(z)}_i q_{t-i}\pmod p
 \text{ for }t=n-k+1,\ldots,n\right\}.
 \label{eq:theory:compatible}
\end{equation}
Then $\mathcal{C}_n$ is nonempty, and every $z\in\mathcal{C}_n$ yields
the same correct next value $q_{n+1}$.
\end{theorycorollary}
\begin{theoryproof}
The generating rule belongs to $\mathcal{C}_n$. Initialize any candidate
$z\in\mathcal{C}_n$ with the $k$ values $q_{n-2k+1},\ldots,q_{n-k}$.
Its next $k$ generated values agree with $q_{n-k+1},\ldots,q_n$ by the
defining checks. The candidate trajectory and the true trajectory
starting at position $n-2k+1$ therefore agree for their first $2k$ values.
Lemma~\ref{lem:theory:continuation} implies that their next values agree.
\end{theoryproof}

The rule itself need not be identifiable: the all-zero trajectory is
compatible with every rule. The construction retains all compatible
rules and uses their common continuation. This distinction is necessary
for a guarantee uniform over initial states.

\subsection{Explicit feed-forward circuits}
\label{app:theory:circuits}
We use three scalar piecewise-linear functions:
\begin{align}
 \tau(u)&=\threlu(u+1)-2\threlu(u)+\threlu(u-1),
 \label{eq:theory:tent}\\
 g(u)&=1-\sum_{b=0}^{J}\tau(u-bp),
 \label{eq:theory:violation}\\
 F(u)&=\threlu(u)-p\sum_{b=1}^{J}
       \bigl[\threlu(u-bp+1)-\threlu(u-bp)\bigr].
 \label{eq:theory:modulo}
\end{align}
Empty sums are zero, including when $J=0$.

\begin{theorylemma}[Modular arithmetic and stability]
\label{lem:theory:circuits}
For every integer $u\in[-B,U]$,
$g(u)=\thind[u\not\equiv0\pmod p]$. On the entire real line,
$0\le g\le1$ and $\thlip(g)\le1$. For every integer $u\in[0,U]$,
$F(u)=u\bmod p$, with the remainder in $\{0,\ldots,p-1\}$.
Furthermore, $F(u)=0$ for $u\le0$ and $\thlip(F)\le B$ globally.
The functions $g$ and $F$ require $3(J+1)$ and $1+2J$ hidden ReLU
units, respectively.
\end{theorylemma}
\begin{theoryproof}
The function $\tau(u)=\max\{1-|u|,0\}$ equals one at the integer zero
and zero at every other integer. The multiples of $p$ in $[-B,U]$ are
precisely $0,p,\ldots,Jp$, which proves the integer formula for $g$.
Since $p\ge2$, the positive supports of the tents in
(\ref{eq:theory:violation}) are disjoint. Their sum lies in $[0,1]$
and has slope in $\{-1,0,1\}$ on each linear piece. This proves the
global range and Lipschitz bounds for $g$.

For integer $u\ge0$,
\[
 \threlu(u-bp+1)-\threlu(u-bp)=\thind[u\ge bp].
\]
For $u\le U$, summing over $1\le b\le J$ gives
$F(u)=u-p\lfloor u/p\rfloor$. All terms vanish when $u\le0$.
The descending intervals $(bp-1,bp)$ are pairwise disjoint, so the
only slopes of $F$ are $0$, $1$, and $1-p$. Their absolute values
are at most $B$. The displayed formulas give the stated unit counts
directly, with an output bias for the constant in $g$.
\end{theoryproof}

\begin{theorylemma}[Candidate selection and linear readout]
\label{lem:theory:readout}
Suppose $E_z\ge0$ and $u_z\in[0,U]$ are integers for
$z=1,\ldots,K$, at least one $E_z$ is zero, and all candidates with
$E_z=0$ satisfy $u_z\bmod p=y$. Set
\begin{align}
 \xi_z&=u_z-(U+1)E_z,\qquad 
 S=\sum_{z=1}^K F(\xi_z),\qquad 
 C=\sum_{z=1}^K\threlu(1-E_z),
 \label{eq:theory:selection}\\
 \lambda_v&=2vS-v^2C,\qquad v=0,\ldots,p-1.
 \label{eq:theory:readout}
\end{align}
Then $S=Cy$, $C\ge1$, and $\lambda_y-\lambda_v=C(y-v)^2\ge1$
for every $v\ne y$. If the $u_z,E_z$ are affine functions of the
FFN input, a single-hidden-layer FFN of width $2K(J+1)$ computes
$(S,C)$.
\end{theorylemma}
\begin{theoryproof}
When $E_z=0$, the contributions to $(S,C)$ are $(y,1)$. When
$E_z\ge1$, $\xi_z\le U-(U+1)=-1$, and both contributions vanish.
Thus $S=Cy$ with $C\ge1$. Substitution into
(\ref{eq:theory:readout}) gives the claimed logit difference. For
each candidate, $F(\xi_z)$ uses $1+2J$ ReLU units and
$\threlu(1-E_z)$ uses one more. All preactivations are affine in the
FFN input, so no additional hidden layer is needed. The FFN output
map forms the two sums directly.
\end{theoryproof}

\subsection{RoPE heads for fixed relative offsets}
\label{app:theory:offsets}

\begin{theorylemma}[Finite-softmax offset retrieval]
\label{lem:theory:offsets}
For a fixed maximum context length $N$, define
\begin{equation}
 \Delta_N=\min_{\substack{u\in\mathbb{Z}\\1\le|u|\le N}}
                    (1-\cos u)>0,
 \qquad \eta_\beta=N\exp(-\beta\Delta_N).
 \label{eq:theory:gap}
\end{equation}
For any offset $i\in\{1,\ldots,k\}$ and $\beta>0$, a head can have
content-independent scores
\begin{equation}
 A_{t,s}=\beta\cos(t-s-i).
 \label{eq:theory:cosine-score}
\end{equation}
If $t-i\ge1$ and $t\le N$, its attention mass at $t-i$ is at least
$1-\eta_\beta$. If each value coordinate lies in an interval of
length $D$, then
\begin{equation}
 \|o_t-\mathbf{V}_{t-i}\|_\infty\le D\eta_\beta.
 \label{eq:theory:copy-error}
\end{equation}
The comparison is to the actual value vector supplied to the head at
position $t-i$.
\end{theorylemma}
\begin{theoryproof}
Using only the first two RoPE coordinates, set
\[
 c_\beta=(\beta\sqrt{d_h})^{1/2},\qquad
 \mathbf{Q}=c_\beta\throt(-i)(1,0)^{\mathsf T},\qquad
 \mathbf{K}=c_\beta(1,0)^{\mathsf T},
\]
with zeros in all remaining coordinates. The shared constant in
(\ref{eq:theory:embedding}) generates these vectors through linear
query and key maps. It will be preserved in every residual update.
Since the first RoPE frequency is one, their scaled inner product is
(\ref{eq:theory:cosine-score}). The target has score $\beta$.
Every other position has a nonzero integer displacement from the
target of absolute value at most $N$. No nonzero integer is a multiple
of $2\pi$, so the finite minimum $\Delta_N$ is positive and every
other score is at most $\beta(1-\Delta_N)$. Therefore
\[
 1-\alpha_{t,t-i}
 \le\sum_{s\ne t-i}\exp(A_{t,s}-\beta)
 \le N\exp(-\beta\Delta_N)=\eta_\beta.
\]
Each coordinate of $o_t-\mathbf{V}_{t-i}$ is a weighted sum of
differences of magnitude at most $D$, with total weight at most
$\eta_\beta$. This proves (\ref{eq:theory:copy-error}).
\end{theoryproof}

If $t-i<1$, the head still returns a well-defined convex combination
of the available values. We will use the retrieval guarantee only
when the target exists, and the convex-hull bound at all other positions.

\subsection{Construction of the two layers}
\label{app:theory:construction}
All active heads use the score in (\ref{eq:theory:cosine-score}), with
the same scale $\beta$. We first analyze an auxiliary computation in
which a head copies its target exactly whenever that target exists.
This is only an intermediate step in the proof. The actual network
uses finite-softmax attention, whose errors are bounded in
Section~\ref{app:theory:softmax}.

\paragraph{Layer 1: local recurrence checks.}
Head $i$, for $i=1,\ldots,k$, selects the original scalar-value
coordinate at offset $i$. The attention output map writes these
values into $k$ separate unused residual coordinates. Denote them by
$w_{t,1},\ldots,w_{t,k}$, and let $w_{t,0}=q_t$, which is preserved
by the residual connection. At $t\ge k+1$, exact retrieval gives
$w_{t,i}=q_{t-i}$ for all $0\le i\le k$. The first FFN writes
\begin{equation}
 e_{t,z}=g\!\left(\sum_{i=1}^k a^{(z)}_i w_{t,i}-q_t\right),
 \qquad z=1,\ldots,K,
 \label{eq:theory:local-checks}
\end{equation}
into $K$ fresh coordinates. For $t\ge k+1$, these are the binary
indicators that rule $z$ violates the observed transition. The affine
argument lies in $[-B,U]$, and Lemma~\ref{lem:theory:circuits} applies.
Its implementation uses $3K(J+1)$ hidden units; each coefficient
$a_i^{(z)}$ is a weight in the first affine map, and each integer
offset $bp$ is a bias.

At earlier positions, some retrieved values do not correspond to
the requested lags. Equation~(\ref{eq:theory:local-checks}) is
nevertheless applied unchanged. These early outputs need not represent
valid checks. At every position, and also for the actual softmax
network, $w_{t,i}\in[0,B]$ and $e_{t,z}\in[0,1]$.

\paragraph{Layer 2: screening and next-value computation.}
For $j=1,\ldots,k-1$, head $j$ selects the entire $K$-dimensional
indicator vector at offset $j$ and writes it to its own fresh group
of residual coordinates. The remaining head has zero value projection.
The residual connection retains the current indicator vector and the
values $w_{t,0},\ldots,w_{t,k}$. At the final position $n$, the second
FFN forms, directly in its affine preactivations,
\begin{equation}
 E_z=\sum_{j=0}^{k-1}e_{n-j,z},\qquad
 u_z=\sum_{i=1}^{k}a_i^{(z)}w_{n,i-1}.
 \label{eq:theory:aggregate}
\end{equation}
The indicator vectors with $j\ge1$ in this formula denote the
retrieved vectors. In the auxiliary exact computation,
\[
 n-j\ge n-k+1\ge k+1\qquad(0\le j\le k-1),
\]
so every check being used is valid and $E_z=0$ precisely when
$z\in\mathcal{C}_n$. Also
$u_z=\sum_{i=1}^k a_i^{(z)}q_{n+1-i}\in[0,U]$.
Corollary~\ref{cor:theory:screening} gives a nonempty set of surviving
candidates, all with $u_z\bmod p=q_{n+1}$.

The FFN implements (\ref{eq:theory:selection}) and writes only the
two statistics $S,C$ into fresh coordinates. In particular, $E_z$,
$u_z$, and $\xi_z$ need no separate residual coordinates or extra
layers. Lemma~\ref{lem:theory:readout} gives hidden width
$2K(J+1)$. The linear output head has entries $2v$ and $-v^2$ on
the $S$ and $C$ coordinates, respectively, and zero elsewhere.
The auxiliary exact computation therefore has correct-class margin
at least one.

\paragraph{Residual width and value dimensions.}
The construction uses the following disjoint residual coordinates:
\begin{center}
\begin{tabular}{lr}
\hline
Stored quantity & Number of coordinates\\
\hline
Shared constant & $1$\\
Original value and $k$ retrieved values & $k+1$\\
Current indicators and $k-1$ retrieved indicator vectors & $kK$\\
Output statistics $S,C$ & $2$\\
\hline
Total & $kK+k+4$\\
\hline
\end{tabular}
\end{center}
Each sublayer writes only into its reserved, initially zero
coordinates. In particular, it preserves the constant and original
value. Rounding the total up to a multiple of $2k$ gives precisely
the value of $d_{\mathrm{model}}$ in (\ref{eq:theory:resources}).
With $H=k$, the head dimension $d_h=d_{\mathrm{model}}/k$ is even
and at least $\max\{2,K\}$. Thus the first layer can use a
two-coordinate RoPE pair, and each active second-layer head can
transport all $K$ indicators within its value dimension. This value
dimension is included in the width bound. The FFN widths are
$3K(J+1)$ and $2K(J+1)$; a common FFN width can be obtained by
adding unused units. When $k=1$, the second layer has no active
attention head; the same FFN and residual construction still applies.

\subsection{Correctness for finite-softmax attention}
\label{app:theory:softmax}
We now prove that the actual network, with the explicit ReLU functions
above and finite attention parameters, preserves the classification
margin. A superscript $\star$ denotes an ideal quantity from the
auxiliary exact computation at positions where its lagged inputs exist;
a hat denotes the corresponding actual quantity. Set
\begin{equation}
 D_E=k(U+1),\qquad D_\xi=U+(U+1)D_E,\qquad
 M=KB^2(2D_\xi+D_E).
 \label{eq:theory:error-constants}
\end{equation}

\paragraph{First-layer error.}
At each $t\ge k+1$, Lemma~\ref{lem:theory:offsets} gives
$|\widehat w_{t,i}-q_{t-i}|\le B\eta_\beta$. The original $q_t$
coordinate is exact. Hence the error in the affine argument of $g$
in (\ref{eq:theory:local-checks}) is at most
\[
 \sum_{i=1}^k a_i^{(z)}B\eta_\beta
 \le kB^2\eta_\beta=U\eta_\beta.
\]
Since $g$ is $1$-Lipschitz,
$|\widehat e_{t,z}-e_{t,z}^\star|\le U\eta_\beta$ at every such
position. At every position, including the early positions,
$\widehat e_{t,z}\in[0,1]$.

\paragraph{Second-layer error.}
Each target position $n-j$, for $1\le j\le k-1$, is at least
$k+1$. The error of retrieving its actual indicator vector is at
most $\eta_\beta$, because all source indicator coordinates lie in
$[0,1]$. Its actual vector is itself within $U\eta_\beta$ of the
ideal vector. Including the current indicator, which does not require
another attention operation, gives
\begin{align}
 |\widehat E_z-E_z^\star|
 &\le\bigl[kU+(k-1)\bigr]\eta_\beta
 \le D_E\eta_\beta,
 \label{eq:theory:check-error}\\
 |\widehat u_z-u_z^\star|
 &\le U\eta_\beta.
 \label{eq:theory:value-error}
\end{align}
The second inequality uses the preserved first-layer values in
(\ref{eq:theory:aggregate}). The bound in
(\ref{eq:theory:check-error}) includes all attention leakage to
early, semantically invalid checks: their range is bounded even
though their interpretation as rule violations need not be valid.

\paragraph{Output error.}
Equations (\ref{eq:theory:selection}),
(\ref{eq:theory:check-error}), and (\ref{eq:theory:value-error}) imply
\[
 |\widehat\xi_z-\xi_z^\star|\le D_\xi\eta_\beta.
\]
Using the global Lipschitz bounds for $F$ and ReLU gives
\[
 |\widehat S-S^\star|\le KB D_\xi\eta_\beta,
 \qquad
 |\widehat C-C^\star|\le K D_E\eta_\beta.
\]
Therefore, for every value token $0\le v\le B$,
\begin{equation}
 |\widehat\lambda_v-\lambda_v^\star|
 \le2v|\widehat S-S^\star|+v^2|\widehat C-C^\star|
 \le M\eta_\beta.
 \label{eq:theory:logit-error}
\end{equation}
This calculation does not require the actual indicator sums or
selection gates to be integers.

\begin{theoryproof}[Completion of Theorem~\ref{thm:theory:two-layer}]
Choose the common attention scale to satisfy
\begin{equation}
 \beta\ge\frac{\log(4NM)}{\Delta_N}.
 \label{eq:theory:finite-scale}
\end{equation}
Then $M\eta_\beta\le1/4$. For the correct value $y=q_{n+1}$,
the auxiliary margin is at least one, so
\[
 \widehat\lambda_y-\widehat\lambda_v
 \ge(\lambda_y^\star-\lambda_v^\star)
       -|\widehat\lambda_y-\lambda_y^\star|
       -|\widehat\lambda_v-\lambda_v^\star|
 \ge1-\tfrac14-\tfrac14=\tfrac12
 \qquad(v\ne y).
\]
All constants depend only on $p,k,K,N$, while the candidate
coefficients are stored as fixed weights. Since $\Delta_N>0$ and
$M<\infty$, the scale and all network parameters are finite.
The same parameters work simultaneously for every legal prefix of
every length $2k\le n\le N$. The residual allocation and FFN counts
in Section~\ref{app:theory:construction} prove
(\ref{eq:theory:resources}), completing the theorem.
\end{theoryproof}

\subsection{Replacing ReLU with GELU}
\label{app:theory:gelu}
The activation assumption in Theorem~\ref{thm:theory:two-layer} can
be relaxed to GELU without increasing depth, head count, residual
width, or FFN hidden width. The modification rescales the FFN
parameters and replaces exact intermediate arithmetic with uniformly
accurate approximations. The positive logit margin then preserves
exact next-token prediction. Throughout this subsection, GELU means
the function
\begin{equation}
 G(x)=x\Phi(x),\qquad
 \Phi(x)=\frac{1}{\sqrt{2\pi}}\int_{-\infty}^{x}e^{-u^2/2}\,du.
 \label{eq:theory:gelu-definition}
\end{equation}

\begin{theorylemma}[Uniform approximation by scaled GELU]
\label{lem:theory:gelu-approximation}
For $\gamma>0$, define $G_\gamma(x)=G(\gamma x)/\gamma$ and
$\varepsilon_\gamma=1/(\gamma\sqrt{2\pi})$. Then
\begin{equation}
 \sup_{x\in\mathbb{R}}
 |G_\gamma(x)-\threlu(x)|\le\varepsilon_\gamma.
 \label{eq:theory:gelu-uniform}
\end{equation}
Replacing the parameters of an FFN by
\begin{equation}
 (W_1,b_1,W_2,b_2)\longmapsto
 (\gamma W_1,\gamma b_1,W_2/\gamma,b_2)
 \label{eq:theory:gelu-rescaling}
\end{equation}
and using activation $G$ realizes the original FFN with every ReLU
replaced by $G_\gamma$, with the same matrix dimensions.
\end{theorylemma}
\begin{theoryproof}
Symmetry of the Gaussian distribution gives
\[
 |G_\gamma(x)-\threlu(x)|=|x|\Phi(-\gamma|x|).
\]
For $u>0$, writing $\phi(v)=e^{-v^2/2}/\sqrt{2\pi}$,
\[
 u\Phi(-u)=\int_u^\infty u\phi(v)\,dv
 \le\int_u^\infty v\phi(v)\,dv
 =\phi(u)\le\frac{1}{\sqrt{2\pi}}.
\]
Substitute $u=\gamma|x|$; at $x=0$, the difference is zero.
This proves (\ref{eq:theory:gelu-uniform}). The rescaled FFN is
\[
 \frac{W_2}{\gamma}G\bigl(\gamma W_1h+\gamma b_1\bigr)+b_2
 =W_2G_\gamma(W_1h+b_1)+b_2,
\]
which proves the implementation claim.
\end{theoryproof}

Let $g_\gamma$ and $F_\gamma$ be the functions in
(\ref{eq:theory:violation}) and (\ref{eq:theory:modulo}), respectively,
with each ReLU replaced by $G_\gamma$. The sums of the absolute
coefficients of their ReLU terms give the global bounds
\begin{equation}
 \begin{aligned}
 \sup_x|g_\gamma(x)-g(x)|&\le A_g\varepsilon_\gamma,
 & A_g&=4(J+1),\\
 \sup_x|F_\gamma(x)-F(x)|&\le A_F\varepsilon_\gamma,
 & A_F&=1+2pJ.
 \end{aligned}
 \label{eq:theory:gelu-circuit-errors}
\end{equation}
Define the finite error constant
\begin{equation}
 \Gamma=K\bigl[B^2kA_g(2U+3)+2BA_F+B^2\bigr],
 \label{eq:theory:gelu-error-constant}
\end{equation}
where $B,U,J$ are from (\ref{eq:theory:arithmetic-constants}).

\begin{theorycorollary}[GELU with unchanged architecture sizes]
\label{cor:theory:gelu}
The conclusion of Theorem~\ref{thm:theory:two-layer}, including the
resource bounds (\ref{eq:theory:resources}) and margin
(\ref{eq:theory:final-margin}), holds with GELU in place of ReLU.
Specifically, use the same construction and rescale both FFNs as in
(\ref{eq:theory:gelu-rescaling}), with
\begin{equation}
 \beta\ge\frac{\log(8NM)}{\Delta_N},\qquad
 \gamma\ge\frac{8\Gamma}{\sqrt{2\pi}}.
 \label{eq:theory:gelu-scales}
\end{equation}
All other architectural assumptions in
Section~\ref{app:theory:model} are retained.
\end{theorycorollary}
\begin{theoryproof}
Compare the GELU network to the ReLU network using the same finite
attention scale $\beta$. Superscripts $\mathrm{G}$ and $\mathrm{R}$
denote these two networks, respectively. This comparison is separate
from the ideal-retrieval comparison in
Section~\ref{app:theory:softmax}.

The first attention layer is identical in both networks, so its
retrieved scalar values agree exactly. The first FFN writes only
the indicator coordinates. By (\ref{eq:theory:gelu-circuit-errors}),
\begin{equation}
 |e^{\mathrm{G}}_{t,z}-e^{\mathrm{R}}_{t,z}|
 \le A_g\varepsilon_\gamma
 \qquad\text{at every position }t\text{ and for every candidate }z.
 \label{eq:theory:gelu-indicator-error}
\end{equation}
This bound includes early positions whose outputs need not be valid
recurrence checks. Both FFNs preserve the shared constant coordinate
and all scalar-value coordinates. Since the second-layer queries and
keys depend only on that constant, its attention weights are also
identical in the two networks. Taking a convex combination cannot
increase the maximum coordinate error. Thus the aggregate and
candidate-value quantities from (\ref{eq:theory:aggregate}) satisfy
\begin{equation}
 |E_z^{\mathrm{G}}-E_z^{\mathrm{R}}|
 \le kA_g\varepsilon_\gamma,\qquad
 u_z^{\mathrm{G}}=u_z^{\mathrm{R}},\qquad
 |\xi_z^{\mathrm{G}}-\xi_z^{\mathrm{R}}|
 \le(U+1)kA_g\varepsilon_\gamma.
 \label{eq:theory:gelu-aggregate-error}
\end{equation}

For the second FFN, compare $F_\gamma(\xi_z^{\mathrm{G}})$ first to
$F(\xi_z^{\mathrm{G}})$ and then to $F(\xi_z^{\mathrm{R}})$.
Use (\ref{eq:theory:gelu-circuit-errors}) for the first difference and
the global Lipschitz bound $\thlip(F)\le B$ for the second.
Applying the analogous argument to
$G_\gamma(1-E_z^{\mathrm{G}})$ and $\threlu(1-E_z^{\mathrm{R}})$
gives
\begin{align}
 |S^{\mathrm{G}}-S^{\mathrm{R}}|
 &\le K\bigl[B(U+1)kA_g+A_F\bigr]\varepsilon_\gamma,
 \label{eq:theory:gelu-S-error}\\
 |C^{\mathrm{G}}-C^{\mathrm{R}}|
 &\le K(kA_g+1)\varepsilon_\gamma.
 \label{eq:theory:gelu-C-error}
\end{align}
The linear readout, with $0\le v\le B$, therefore satisfies
\begin{equation}
 |\lambda_v^{\mathrm{G}}-\lambda_v^{\mathrm{R}}|
 \le 2B|S^{\mathrm{G}}-S^{\mathrm{R}}|
       +B^2|C^{\mathrm{G}}-C^{\mathrm{R}}|
 \le\Gamma\varepsilon_\gamma.
 \label{eq:theory:gelu-logit-error}
\end{equation}
In particular, we do not assume that the GELU indicators lie in
$[0,1]$ or remain exactly binary. Their uniform closeness to the ReLU
indicators suffices.

By (\ref{eq:theory:logit-error}), the ReLU logits differ from the
ideal logits by at most $M\eta_\beta$. The scale choices in
(\ref{eq:theory:gelu-scales}) ensure
$M\eta_\beta\le1/8$ and $\Gamma\varepsilon_\gamma\le1/8$.
Consequently, every GELU logit is within $1/4$ of its ideal value.
The ideal margin is at least one, so for $y=q_{n+1}$ and $v\ne y$,
\[
 \lambda_y^{\mathrm{G}}-\lambda_v^{\mathrm{G}}
 \ge1-2\bigl(M\eta_\beta+\Gamma\varepsilon_\gamma\bigr)
 \ge\tfrac12.
\]
The approximation bounds are uniform over real inputs, so one finite
choice of $\beta,\gamma$ works for every legal prefix with
$2k\le n\le N$. Rescaling changes no matrix dimensions or reserved
residual coordinates, which proves the same resource bounds.
\end{theoryproof}

